\pdfoutput=1

\documentclass[11pt]{article}

\usepackage[preprint]{acl}

\usepackage{inconsolata}
\usepackage{times}
\usepackage{diagbox}
\usepackage{booktabs}
\usepackage{multirow}
\usepackage{tablefootnote}
\usepackage{array}
\usepackage{pdfpages}
\usepackage{colortbl}

\usepackage{amsmath}
\usepackage{amsfonts}
\usepackage{pifont}
\usepackage{makecell}
\usepackage{longtable}
\usepackage{geometry}
\usepackage[most]{tcolorbox}
\usepackage{graphicx}
\usepackage{multicol}

\usepackage{float}
\usepackage{subcaption}
\usepackage{paralist}
\usepackage{enumitem}
\usepackage{algorithm}

\usepackage{algorithmic}
\definecolor{lightgreen}{RGB}{145, 204, 117}
\definecolor{lightyellow}{RGB}{250, 200, 88}
\definecolor{lightred}{RGB}{238, 102, 102}
\definecolor{lightblue}{RGB}{115, 192, 222}
\definecolor{lightorange}{RGB}{253, 208, 162} 

\newtcolorbox{promptbox}[3][Judge Prompt]{
colback=black!5!white,
arc=5pt, 
boxrule=0.5pt,
fonttitle=\bfseries,
title=#1, 
before upper={\small}, fontupper=\fontfamily{ptm}\selectfont,
colframe=#2,
label=#3,
}

\usepackage{titlesec}

\usepackage[T1]{fontenc}

\newcommand{\vic}[1]{\textcolor{black}{#1}}

\usepackage{amssymb}

\usepackage[utf8]{inputenc}

\usepackage{microtype}

\usepackage{inconsolata}

\usepackage{graphicx}

\title{From Sufficiency to Reflection: Reinforcement-Guided Thinking Quality in Retrieval-Augmented Reasoning for LLMs	}

\author{
    Jie He$^1$ \quad 
    \textbf{Víctor Gutiérrez Basulto}$^2$\footnotemark[1] \quad and \; 
    \textbf{Jeff Z. Pan}$^1$\thanks{ \, Corresponding author}  \\
     $^1$ School of Informatics, University of Edinburgh, UK  \\
    $^2$ School of Computer Science and Informatics, Cardiff University, UK \\
    \normalsize{\texttt{ \{j.he, j.z.pan\}@ed.ac.uk }} \normalsize{\texttt{gutierrezbasultov@cardiff.ac.uk}} \\
}

\begin{document}
\maketitle
\begin{abstract}

\vic{Reinforcement learning-based retrieval-augmented generation (RAG) methods enhance the reasoning abilities of large language models (LLMs). However, most rely only on final-answer rewards, overlooking intermediate reasoning quality. This paper analyzes existing RAG reasoning models and identifies three main failure patterns: (1) \emph{information insufficiency:} failure to retrieve adequate support; (2) \emph{faulty reasoning:} logical or content-level flaws despite sufficient information; (3) \emph{answer–reasoning inconsistency:} a valid reasoning chain with a mismatched final answer. We propose \textbf{TIRESRAG-R1}, a novel framework using a \textit{think–retrieve–reflect} process and a multi-dimensional reward system to improve reasoning and stability. TIRESRAG-R1 introduces: (1) a \emph{sufficiency reward} to encourage thorough retrieval; (2) a \emph{reasoning quality reward} to assess rationality and accuracy of the reasoning chain; (3) a \emph{reflection reward} to detect and revise errors. It also employs a \emph{difficulty-aware reweighting strategy} and \emph{training sample filtering} to boost performance on complex tasks. Experiments on four multi-hop QA datasets show TIRESRAG-R1  outperforms prior RAG methods and generalizes well to single-hop tasks\footnote{The code and data can be found at \href{https://github.com/probe2/TIRESRAG-R1}{https://github.com/probe2/TIRESRAG-R1}.}}.

\end{abstract}
\section{Introduction}


\vic{Large language models (LLMs)~\cite{grattafiori2024llama3herdmodels, yang2025qwen3technicalreport, openai2024gpt4ocard} have achieved remarkable performance across a wide range of downstream tasks, such as 
mathematical reasoning \cite{ahn-etal-2024-large}, 
code generation \cite{10.1145/3747588}, open-domain question answering \cite{kamalloo-etal-2023-evaluating}.
A key factor behind this success is \emph{chain-of-thought prompting}~\cite{10.5555/3600270.3602070}, which enables LLMs to generate intermediate reasoning steps before arriving at a final answer, significantly enhancing performance on reasoning tasks~\cite{wang-etal-2023-plan,pan-etal-2023-logic,snell2025scaling}. Despite these advances, the internal knowledge of LLMs is not always reliable. For instance, when faced with time-sensitive queries~\cite{mousavi2024dyknowdynamicallyverifyingtimesensitive} or conflicting evidence within their internal representations~\cite{xu-etal-2024-knowledge-conflicts}, LLMs often produce hallucinations or factual inaccuracies due to outdated or ambiguous information~\cite{li2024enhancingllmfactualaccuracy, wang2024resolving}.}


\begin{figure}
    \centering
    \includegraphics[width=1\linewidth]{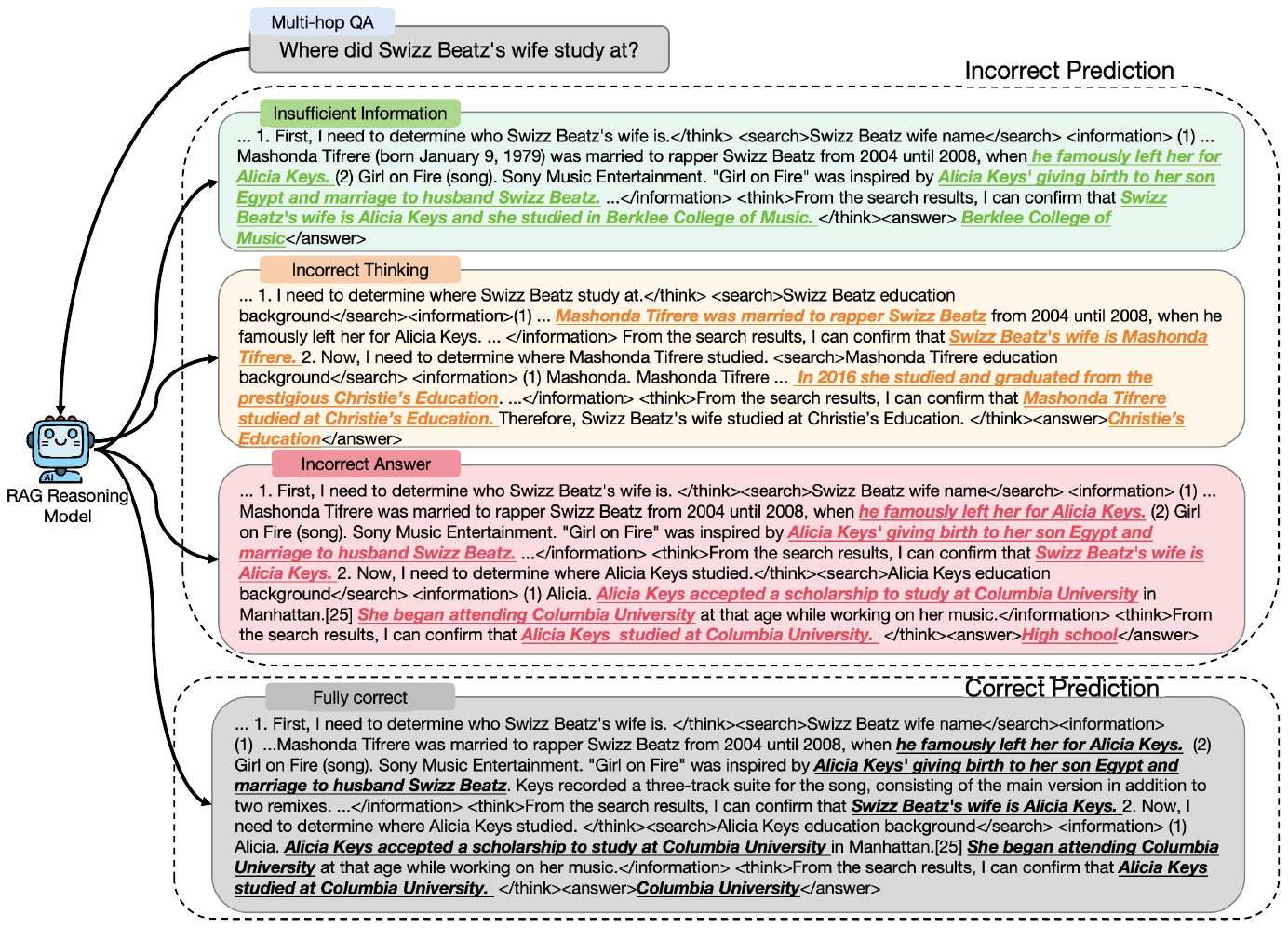}
    \caption{\vic{An example showing different reasoning trajectories for answering a multi-hop  query. It compares insufficient information, incorrect predictions, and fully correct reasoning.}}
    \label{fig:intro}
\end{figure}

\vic{The \emph{retrieval-augmented generation (RAG)} paradigm~\cite{gao_retrieval-augmented_2024} improves factual accuracy and robustness by enabling LLMs to access up‑to‑date external knowledge. However, standard RAG often treats retrieval and generation as loosely coupled, lacking mechanisms for multi‑step reasoning such as query decomposition or identifying knowledge gaps, which limits performance on tasks requiring deeper reasoning.}

\vic{To overcome this, a growing body of research has turned to \emph{reinforcement learning (RL)} (a paradigm that has demonstrated strong performance in mathematical reasoning and code generation~\cite{deepseek-math, wang2024rlcoderreinforcementlearningrepositorylevel}) to train LLMs for retrieval-augmented reasoning.
In this setting, the model learns when and how to retrieve, as well as how to integrate retrieved information into coherent reasoning chains, guided by outcome-level rewards (e.g., answer correctness)~\cite{jin2025searchr1trainingllmsreason, song2025r1searcherincentivizingsearchcapability, sun2025zerosearchincentivizesearchcapability}.}
\vic{Despite recent progress, outcome-based RL methods still face two significant limitations: (1) they often neglect the quality and validity of intermediate reasoning steps, and (2) they lack fine-grained feedback to guide the reasoning process during multi-step tasks. As a result, models may learn incorrect reasoning paths that degrade answer quality, or they may arrive at correct answers through flawed reasoning, thereby compromising interpretability.}

\vic{In this paper, to address these challenges, we begin by evaluating the performance of RAG-based reasoning models trained with outcome-based rewards. Through human analysis on examples from multi-hop reasoning datasets (e.g., MuSiQue \cite{trivedi-etal-2022-musique})%
, as shown in Fig.~\ref{fig:intro}, we identify three primary bottlenecks limiting model accuracy: (1) the model retrieves sufficient information but still fails to produce the correct answer; (2) reasoning is prematurely interrupted, resulting in inadequate retrieval; and (3) the reasoning chain is correct, yet the final answer is incorrect. Based on these insights, we propose \textbf{TIRESRAG-R1}, an approach that enhances reasoning chain quality by incorporating rewards focused on sufficiency and reasoning coherence. Additionally, we introduce a \emph{thinking–retrieval–reflection} framework, motivated by the principle that models should first ensure the integrity of their reasoning chains. If the final answer is incorrect despite a valid chain, the model engages in reflection to revise its response.}

\vic{
TIRESRAG-R1 enhances RAG-based reasoning by introducing a thinking–retrieval–reflection framework, where the model can decide post-answer whether to reflect and revise its output through an additional round of reasoning and retrieval. To guide learning, TIRESRAG-R1 assigns multi-dimensional reward signals to each reasoning trajectory, including: an answer reward, a sufficiency reward to encourage retrieval of adequate evidence, a reasoning quality reward evaluating logical coherence, alignment, error awareness, and conciseness, and a reflection reward that promotes correction of wrong answers while discouraging unnecessary changes. To address the varying difficulty of retrieval tasks, we propose a difficulty-aware advantage reweighting strategy that adjusts reward weights based on the ease of retrieving sufficient evidence, and an advantage filtering mechanism that excludes trivial examples (where all responses are correct or incorrect) to stabilize RL training. Together, these components enable more robust and interpretable reasoning, significantly outperforming traditional outcome-based methods across multiple QA benchmarks.}


\vic{Our contributions  can be summarized as follows:}
\vic{\begin{itemize}
\setlength{\itemsep}{0pt}
\setlength{\parsep}{0pt}
\setlength{\parskip}{0pt}
  \item We are the first to define \textit{overthinking} and \textit{underthinking} in RAG and conduct comprehensive analysis to reveal the main bottlenecks faced by current RL-trained RAG reasoning models.
  \item  We propose \textbf{TIRESRAG-R1}, which enhances reasoning by introducing a reflection mechanism and fine-grained multi-dimensional rewards to improve both reasoning chains and answer accuracy. We also propose a difficulty-aware advantage reweighting strategy with no additional computational cost to guide the model toward optimizing on harder questions.
  \item Our experiments show that TIRESRAG-R1 outperforms multiple state-of-the-art RAG methods on several open-ended QA datasets, with an average accuracy improvement of 5.8\%. Further ablation studies and analysis confirm the effectiveness of our fine-grained rewards.
\end{itemize}}

\section{Related Work}
\vic{\noindent \textbf{Retrieval-Augmented Generation. }
RAG has been widely adopted to mitigate hallucination, domain knowledge gaps, and temporal staleness \cite{ayala-bechard-2024-reducing,siriwardhana-etal-2023-improving,gade2024itstimeincorporatingtemporality}. Traditional RAG typically follows a static retrieve‑then‑generate pipeline \cite{NEURIPS2020_6b493230,10.5555/3524938.3525306}, which is effective for open‑domain QA but often struggles with multi‑hop reasoning, latent constraints, and ambiguous queries \cite{tang2024multihoprag,li2024elicitreasoningllmscriticguided,chan2024rqrag}. Recent work has proposed more cognitively inspired architectures, such as AdaptiveRAG \cite{jeong-etal-2024-adaptive} using query classification to choose retrieval strategies. In parallel, modular and hybrid frameworks \cite{gao2024modularragtransformingrag,zhang2024hierarchicalretrievalaugmentedgenerationmodel,zhou2024metacognitive} integrate evidence aggregation, verification, and query rewriting, reflecting a deeper interaction between retrieval and reasoning. }

\noindent \textbf{Reinforcement Learning for LLM Reasoning. }
Reinforcement learning (RL) has recently been explored as a way to improve multi‑step reasoning in LLMs. Models such as GPT‑o1 and DeepSeek‑R1 \cite{openai2024openaio1card,deepseekai2025deepseekr1incentivizingreasoningcapability} demonstrate RL‑based training for structured reasoning. Follow‑up work (e.g., SimpleRL‑Zoo \cite{zeng2025simplerlzooinvestigatingtamingzero}, Open‑Reasoner‑Zero \cite{hu2025openreasonerzeroopensourceapproach}, and Light‑R1 \cite{wen2025lightr1curriculumsftdpo}) further leverage curriculum design, reward shaping, and improved optimization algorithms such as Dr GRPO \cite{liu2025understandingr1zeroliketrainingcritical} to enhance verifiability and coherence.

\smallskip
\noindent \vic{\textbf{Reinforcement Learning for RAG Reasoning. }
Recent studies combine RL with RAG to dynamically guide when and what to retrieve \cite{song2025r1searcherincentivizingsearchcapability,jin2025searchr1trainingllmsreason,chen2025researchlearningreasonsearch}. Early approaches rely primarily on outcome‑level rewards, while newer ones introduce process‑level rewards \cite{wang2025stepsearchignitingllmssearch,sha2025semreinforcementlearningsearchefficient,zhang2025processvsoutcomereward}. For instance, R3‑RAG \cite{li2025r3raglearningstepbystepreasoning} computes document relevance at each step to refine search strategies, and others enrich the “search–think–answer” pipeline with additional fields to prompt deeper reflection \cite{ren2025effectivetransparentragadaptivereward,shi2025searchrefinethinkautonomous}. In contrast, our work directly rewards the model’s reasoning process by measuring the sufficiency of retrieved documents, enabling more adaptive retrieval‑reasoning coordination.} 

For more details on related work, please see Appendix~\ref{sec:rel}.

\section{Prelimary Study}
\label{sec3}
\vic{Although R1-like RAG models demonstrate stronger reasoning capabilities than traditional RAG models in QA tasks, questions remain about the faithfulness of their reasoning processes, both in relation to the provided context and in how well their final answers reflect that reasoning.
This section focuses on the following central question:
\textit{In current reasoning-oriented RAG models tackling multi-hop tasks, what is the correlation between the predicted answers, the retrieved documents, and the generated reasoning text?}}

\begin{table}[!t]
\footnotesize
\centering
\begin{tabular}{lcccc}
\toprule
\diagbox[width=3cm]{Category}{Datasets} & \multicolumn{2}{c}{2Wiki} & \multicolumn{2}{c}{MuSiQue} \\
\hline 
 & Corr. & Incorr. & Corr. & Incorr. \\
\midrule
Overthinking   & 13  & 25  & 13  & 31  \\
Good thinking  & 232 & 89  & 80  & 96  \\
Underthinking  & 16  & 125 & 0   & 280 \\
\bottomrule
\end{tabular}
\caption{\vic{Distribution of correct (Corr.) and incorrect (Incorr.) predictions across different reasoning categories on 2Wiki and MuSiQue datasets.}}
\label{tab:reasoning-stats}
\vspace{-0.2cm
}
\end{table}

\subsection{Preliminary Analysis}  
\noindent \vic{We trained a reasoning‑capable RAG model using the GRPO algorithm from DeepSeek‑R1, following the data setup of R1‑Searcher. We used Qwen‑2.5‑3B‑Instruct with a local retriever and evaluated on in‑domain 2Wikimultihopqa (2Wiki) and out‑of‑domain Musique test sets.}

\vic{In what follows, we will denote an example as $E = (Q, RD, A)$, where $Q$ is the query, $RD$ the combined reasoning steps and retrieved documents, and $A$ the predicted or gold answer.}
\vic{Regarding reasoning behavior, we categorize the model’s reasoning into three cases: \textbf{overthinking} (gold answer could be inferred before the final step), \textbf{good thinking} (gold answer emerges exactly at the final step), and \textbf{underthinking} (model stops without sufficient content). Following prior work \cite{joren2025sufficient}, we use an LLM to assess whether $Q$ and $RD$ are sufficient to derive the gold answer, but unlike \cite{joren2025sufficient}, we also provide the gold answer to reduce misjudgments. Details of fine‑tuning setup and experimental analysis settings are all provided in Appendix~\ref{prompt_tem}.}

\begin{table*}[!htbp]
\scriptsize
\centering
\begin{tabular}{cp{0.2\textwidth}p{0.5\textwidth}c}
\hline
\textbf{Context Sufficiency }&\textbf{Error type} & \textbf{Case} &\textbf{Ratio} \\
\hline
Insufficient&Knowledge retrieved incorrect& \textbf{Question:} What is the place of birth of the founder of Leister Technologies? 
\textbf{Model output:} ... \textcolor{blue}{<search>} founder of Leister Technologies \textcolor{blue}{</search>} ... 
\textcolor{red}{<think>} Louis Leitz (founder of Esselte Leitz) ... therefore answer Ingersheim \textcolor{red}{</think>} \textcolor{orange}{<answer>} Ingersheim \textcolor{orange}{</answer>}  \textbf{Gold Answer:} Wilhelmshaven& 30.2\% \\
\hline
Insufficient&Thinking incorrect & \textbf{Question:} Where was the place of burial of John De Ferrers, 1st Baron Ferrers of Chartley's father? 
\textbf{Model output:} ... \textcolor{blue}{<search>} John De Ferrers, 1st Baron Ferrers of Chartley father's burial place \textcolor{blue}{</search>} ...
\textcolor{red}{<think>} Linked to Sir Humphrey Ferrers of Tamworth Castle and concluded burial at Tamworth Church \textcolor{red}{</think>} \textcolor{orange}{<answer>} Tamworth Church \textcolor{orange}{</answer>}
\textbf{Gold Answer:} Staffordshire
&22.1\% \\
\hline
Sufficient&Thinking process mismatch evidence  &\textbf{Question:} Who is Gaius Rubellius Blandus's father-in-law?  
\textbf{Model output:} ... \textcolor{blue}{<search>} Gaius Rubellius Blandus father-in-law \textcolor{blue}{</search>} ... 
\textcolor{red}{<think>} Married Julia (granddaughter of Tiberius) $\Rightarrow$ therefore father-in-law is Tiberius \textcolor{red}{</think>}\textcolor{orange}{ <answer>} Tiberius \textcolor{orange}{</answer>}  
\textbf{Gold Answer:} Drusus Julius Caesar
&16.4\% \\
\hline
Sufficient&Thinking correct but incorrect answer&\textbf{Question:} Which film has the director died earlier, Girls School Screamers or Lucky Numbers?  
\textbf{Model output:} ... \textcolor{blue}{<search>} director of Girls School Screamers \textcolor{blue}{</search>}  ... \textcolor{blue}{<search>} director of Lucky Numbers \textcolor{blue}{</search>}  ...
\textcolor{red}{<think>} Found John P. Finnegan (died 2007) and Nora Ephron (died 2012) $\Rightarrow$ therefore the director of Lucky Numbers director died earlier \textcolor{red}{</think>} \textcolor{orange}{<answer>}Girls School Screamers\textcolor{orange}{</answer>} 
\textbf{Gold Answer:} Lucky Numbers  
& 5.7\% \\
\hline
Sufficient&Thinking process match evidence but thinking  process is incorrect &\textbf{Question:} Which film has the director who was born first, Nasib Si Labu Labi or Solo Per Te Lucia?  
\textbf{Model output:} ... \textcolor{blue}{<search>} director of Nasib Si Labu Labi birth year \textcolor{blue}{</search>}  ... \textcolor{blue}{<search>} director of Solo Per Te Lucia birth year \textcolor{blue}{</search>}  ...
\textcolor{red}{<think>} P. Ramlee (born 1936) vs Franco Rossi (born 1923) $\Rightarrow$ therefore Nasib Si Labu Labi director was born first \textcolor{red}{</think>}\textcolor{orange}{<answer>} Nasib Si Labu Labi \textcolor{orange}{</answer>}  
\textbf{Gold Answer:} Solo Per Te Lucia
 & 25.6\% \\
\hline
\end{tabular}
\caption{\vic{Fine-grained error analysis with examples.}}
\label{tab:prelimary}
\end{table*}

\vic{We report sufficiency judgment results separately based on whether the model’s prediction is correct. The results are shown in Table~\ref{tab:reasoning-stats}.  
We can observe that among correctly predicted examples, overthinking and underthinking account for very small proportions, while good thinking accounts for 86\% in Musique and 89\% in 2Wiki. This indicates that for correctly predicted examples, the model’s reasoning process is generally just right and sufficient.  
Among incorrectly predicted examples, underthinking accounts for a large proportion. This is unsurprising, as the model did not retrieve relevant documents and thus lacked enough information to answer the question, so answering incorrectly is expected.  
Surprisingly, overthinking and good thinking together account for 48\% in Musique and 31\% in 2Wiki. This shows that the reasoning process was sufficient, but the model still failed to produce the correct answer.} 

\subsection{Fine-grained Analysis}  
\vic{Since overthinking accounts for a small proportion among predictions, we further analyzed the incorrectly predicted examples belonging to good thinking and overthinking to determine specific error causes. We manually reviewed these examples and classified them into five error types, as shown in Table~\ref{tab:prelimary}.}  

\vic{ We observe the following. When the reasoning process was insufficient to fully answer the question, we found two error types:  
(i) the model mistakenly believes the retrieved content matches the question entity and therefore stops further retrieval and outputs an answer, while in fact the retrieved content is irrelevant;  
(ii) the model only obtained partial content relevant to the question and ended the reasoning, which is a reasoning omission.} \vic{ When the reasoning process was sufficient to answer the question, a prominent problem was that the model failed to follow evidence, indicating difficulty in understanding retrieved content and integrating it into reasoning \cite{shi2025searchrefinethinkautonomous,li2025searcho1agenticsearchenhancedlarge}.  
Another issue was that the reasoning process was correct but the final answer was wrong, or the model showed correct evidence but failed to reason correctly from it.}

\vic{Based on these findings, we categorize the failure cases of RAG reasoning on part of the test sets into three major types: \textbf{retrieval failure}, \textbf{reasoning failure}, and \textbf{answer failure}.  Motivated by this observation, we propose \textbf{TIRESRAG-R1}, which 
aims to enhance the model’s awareness of retrieval sufficiency while improving both its reasoning chain and final answer quality.}

\begin{figure*}
    \centering
    \includegraphics[width=1\linewidth]{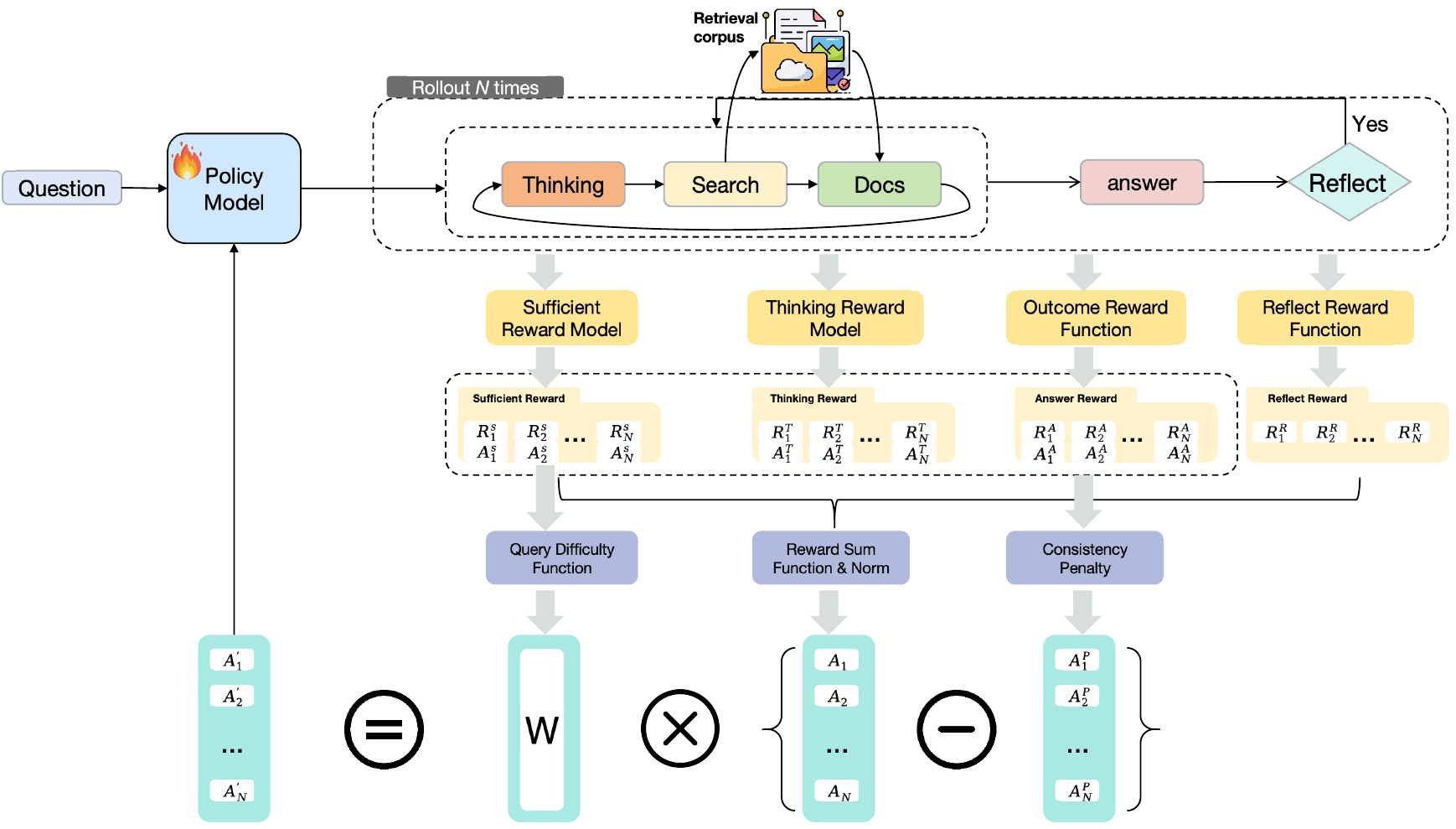}
    \caption{Overall architecture of TIRESRAG-R1, which integrates search, reasoning, and reflection with a dynamic reward design to guide reinforcement learning.}
    \vspace{-0.3cm}

    \label{fig:main_method}
\end{figure*}
\section{Method}
\vic{In this section, we first introduce our GRPO-based \textit{thinking–search–answer–reflect} pipeline.  
Then, we elaborate on the reward design, which is the core of our RL algorithm.  
On top of the standard answer reward, we carefully add three additional reward signals: \textit{thinking reward}, \textit{sufficient reward}, and \textit{reflect reward}.  
Finally, we observe that during training there exists a class of ``extreme'' examples that  are either answered correctly by the model in all rollouts or answered incorrectly in all rollouts.  
Such examples contribute very limited training signals and may even introduce noise, affecting training stability.  To address this issue, we introduce two optimization mechanisms: a \textit{difficulty‑aware advantage reweighting strategy} and a \textit{group filtering mechanism} that filters out those extreme problems. A complete algorithmic description can be found in Appendix~\ref{app_algorithm}.} 

\subsection{Trajectory Generation with Search and  Reflect}
\label{sec:traj}
\vic{Given a question $q$, we first prompt the LLM $\pi_\theta$ to generate a long chain of thought.  
During this thinking process, the model is encouraged to trigger search operations in order to use external document information.  
When the model decides that a search is needed, it terminates the current thinking step. The reasoning text so far is wrapped with \texttt{<think>} \texttt{</think>} tags, and the search query is wrapped with \texttt{<search>} \texttt{</search>} tags.  
We extract the search query and feed it into the retriever $\pi_{\text{ret}}$ to obtain $k$ relevant documents $D_k = \{d_1, \dots, d_k\}$.  
These documents are wrapped with \texttt{<information>} \texttt{</information>} tags and appended back to the trajectory.  
The model then continues reasoning based on the retrieved documents.  
This process of thinking and searching repeats until the model judges that enough information has been collected to generate the final answer.  
At that point, the answer is wrapped with \texttt{<answer>} \texttt{</answer>} tags.  
However, according to our prompt design, the model is instructed to reflect on the generated answer, potentially entering another cycle of thinking and searching before producing a second answer wrapped again with \texttt{<answer>} \texttt{</answer>}.  
This reflection mechanism is specifically introduced to address the issue discussed in Section ~\ref{sec3}, namely that the reasoning process may be correct but the first generated answer is wrong.  
In the final trajectory, the content inside the last \texttt{<answer>} \texttt{</answer>} is taken as the model’s predicted answer.}

\subsection{GRPO for Training}
\vic{As shown in Figure~\ref{fig:main_method}, we adopt the \emph{group relative policy optimization (GRPO)} algorithm \cite{deepseekai2025deepseekr1incentivizingreasoningcapability}  for RL training.
For each query in GRPO, a group of $G$ rollout trajectories, as described in Section~\ref{sec:traj}, is generated using the current policy $\pi_{\theta}^{\text{old}}$.  
Here $\pi_{\theta}^{\text{old}}$ also serves as a frozen reference model initialized with the same parameters as the policy model.  
The GRPO algorithm uses the following optimization objective to update the policy: } 
\begin{equation*}
{
\footnotesize
\begin{aligned}
J_{\text{GRPO}}(\theta) = {} &
\mathbb{E}_{q \sim Q,\{o_i\}_{i=1}^{G} \sim \pi_{\text{old}}} \Bigg[
\frac{1}{G} \sum_{i=1}^{G} \sum_{t=1}^{|o_i|}
\min \\ \Bigg(
\frac{\pi_\theta(o_{i,t}\mid q)}{\pi_{\theta_{\text{old}}}(o_{i,t}\mid q)} A_i, 
& 
\text{clip}\!\left(
\frac{\pi_\theta(o_{i,t}\mid q)}{\pi_{\theta_{\text{old}}}(o_{i,t}\mid q)},
1-\epsilon, 1+\epsilon
\right) A_i
\Bigg)
\Bigg]
\end{aligned}
}
\end{equation*}
\vic{where $\epsilon$ is the clipping hyper‑parameter and $\beta$ is the KL‑divergence penalty coefficient.  
The advantage $A_i$ for each response is computed as 
$A_i^ = \tfrac{ r_i  - \text{mean}(\{r_j\}_{j=1}^{G}) }
{ \text{std}(\{r_j\}_{j=1}^{G}) }$,
where $\{r_i\}_{i=1}^G$ are the rewards from the group.  
The detailed definition of each reward is introduced in Section~\ref{sec:reward}.}

\subsection{Reward Design}
\label{sec:reward}
\vic{As described in Section~\ref{sec3}, we aim to achieve three major goals:  
(1) encourage the policy model to search sufficiently before generating an answer,  
(2) improve the quality of the reasoning chain and make it more consistent with the retrieved information, and  
(3) enhance the model’s ability to correct answers through reflection.  
To achieve these goals, we adopt a mixed reward function with the following components:}




\noindent \textbf{Answer reward }  
\vic{measures the match between the predicted answer and the gold answer.  
Following prior work \cite{jin2025searchr1trainingllmsreason}, we use the F1 score as the answer reward to avoid reward hacking issues observed with exact match (EM):
\begin{equation}
\small
  R^A = F_{1}(a, a^*).
\end{equation}}
\vic{\noindent\textbf{Sufficient reward }  
measures whether the reasoning trajectory $RD$ provides enough information to support the gold answer.  
Following Section~3, we use a locally deployed LLM  to score $(Q, RD, \text{gold answer})$ with 0/1, where 1 means sufficient and 0 means insufficient:
\begin{equation}
\small
  R^S =
  \begin{cases}
    1.0, & \text{if $(Q, RD, o_g)$ is sufficient}, \\
    0.0, & \text{otherwise}.
  \end{cases}
\end{equation}}
\vic{\noindent \textbf{Thinking reward }  
measures the quality of the thinking part of the trajectory.  
The locally deployed LLM is prompted to evaluate logical soundness, alignment with retrieved content, error awareness, and concise accuracy.  
It outputs a score in $[0,1]$:
\begin{equation}
\small
  R^T \in [0, 1].
\end{equation}}
\vic{\noindent \textbf{Reflect reward }  
encourages the model to revise wrong answers into correct ones.  
We extract each intermediate answer $a_1, a_2, \dots$ from the trajectory, compare each with the gold answer using an accuracy score, and assign rewards as follows: a positive reward when reflection corrects an incorrect answer, a negative reward when it replaces a correct answer with an incorrect one, and zero otherwise.}

{\scriptsize
\begin{equation}
  R^R =
  \begin{cases}
    +1.0, & \text{if } CEM(a_1, a^*) = 0  \text{ and }  CEM(a_2, a^*) = 1, \\
    -1.0, & \text{if } CEM(a_1, a^*) = 1 \text{ and } CEM(a_2, a^*) = 0, \\
    0.0, & \text{otherwise}.
  \end{cases}
\end{equation}
}

\noindent \textbf{Dynamic Weight of Reward. }\vic{Because of the mixed reward mechanism, the model could overfit to auxiliary signals such as sufficient reward or thinking reward and ignore answer accuracy. We design a dynamic weight schedule $a_t = \tfrac{1}{1 + \exp\!\big(\tfrac{t - 0.9 T}{10}\big)}$, where $t$ denotes the current training step and $T$ denotes the total number of training steps. This schedule gradually shifts focus from auxiliary reasoning rewards to answer accuracy as training proceeds.}

\smallskip
\vic{Finally, the overall reward is computed as}
\begin{equation}
    R^{\text{sum}} = R^A + a_t \bigl( w_t R^T + w_s R^S + w_r R^R \bigr),
\label{reward_sum}
\end{equation}

\subsection{Optimization Strategy}
\label{sec:4.4}
\textbf{Difficulty‑aware Resampling. }  
\vic{Despite employing a mixed reward mechanism, we observed that different rollout groups can yield very different raw rewards but almost identical normalized advantages (e.g., $[0.8,0.85,0.9,0.95,1.0]$ vs.\ $[0.0,0.05,0.1,0.15,0.2]$). This phenomenon causes simple problems to receive the same optimization emphasis as hard problems. Following \citet{zhang2025grpoleaddifficultyawarereinforcementlearning}, we introduce a difficulty‑aware advantage reweighting strategy. For each sample, we estimate the problem difficulty by computing the average sufficient‑reward score across all its rollouts, denoted as $R_{S}^{\text{avg}}$. The weight function is defined as}
\begin{equation}
\small
    W(R^{S}_{\text{avg}}) = A + \frac{B - A}{1 + \exp\!\big[k(R^{S}_{\text{avg}} - \rho_0)\big]},
\label{diff-sample}
\end{equation}

\vic{where $A, B, \rho_0, k$ are tunable hyper‑parameters controlling the sensitivity of the reweighting. $R^{s}_{\text{avg}}$ is the average sufficient‑reward score of all rollouts associated with question $q$.}

\noindent \textbf{Consistency Penalty. }  
\vic{Because we use a mixed reward mechanism, it is possible for a trajectory to receive a high reasoning reward but a low answer reward.  
To encourage consistent trajectories where high reasoning quality aligns with high answer accuracy, we add a small penalty term to discourage such inconsistencies:
\begin{equation}
\small
A_i^P = -\lambda_p  A_i^{T} \cdot  A_i^{T} \cdot  A_i^{A}, \quad \text{if }  A_i^{S} \cdot  A_i^{T} \cdot A_i^{A} < 0.
\label{con-pen}
\end{equation}
\vic{where $ A_i^{S}$, $A_i^{T}$, and $A_i^{A}$ denote the group‑normalized values of the sufficient reward, the thinking reward, and the answer reward, respectively. The final difficulty‑aware advantage is then expressed inline as $A_i' = (A_i - A_i^P) \cdot W(R^{s}_{\text{avg}})$, where $A_i$ is the original normalized advantage for rollout $i$. 
}}

\noindent \textbf{Group Filtering. }  
\vic{As noted in \cite{jin2025searchr1trainingllmsreason}, applying GRPO to RAG tasks can lead to late‑stage collapse, where training rewards drop to zero.  
We also observe this phenomenon: in later stages, many queries have groups where all rollouts are either completely correct or completely wrong, leading to zero advantage and noisy updates.  
To mitigate this, we introduce a simple filtering mechanism that removes these saturated queries from the training set.} 

\begin{table*}[htbp]
\footnotesize
\begin{tabular}{lcccccccccccc}
\toprule
Method & \multicolumn{3}{c}{Hotpotqa} & \multicolumn{3}{c}{2wikimultihopqa} & \multicolumn{3}{c}{Musique} & \multicolumn{3}{c}{Bamboogle} \\
 & EM & F1 & Judge &  EM & F1 & Judge & EM & F1 & Judge &EM & F1 & Judge \\
\midrule
Direct Generation & 18.0 & 23.5 & 24.2 & 19.0 & 23.6 & 23.0 & 3.6 & 9.0 & 6.0 & 20.8 & 30.7 & 28.8 \\
COT & 18.2 & 24.4 & 25.2 & 21.0 & 25.7 & 25.0 & 4.4 & 10.9 & 8.8 & 21.6 & 31.3 & 28.8 \\
\hline 

Naive RAG & 29.4 & 40.7 & 43.4 & 26.0 & 30.5 & 29.8 & 5.2 & 11.0 & 8.0 & 18.4 & 27.0 & 22.4 \\
Sure & 20.2 & 27.2 & 32.0 & 21.0 & 24.6 & 26.2 & 3.0 & 8.5 & 5.2 & 10.4 & 17.9 & 12.8 \\
IRCOT & 22.2 & 32.7 & 38.6 & 17.4 & 23.4 & 26.0 & 5.2 & 10.8 & 8.4 & 13.6 & 23.8 & 24.0 \\

Self-ask & 14.4 & 24.6 & 39.0 & 14.8 & 19.6 & 25.0 & 4.0 & 10.3 & 7.4 & 9.6 & 22.3 & 18.4 \\
RAG with Agentic Search & 7.0 & 10.1 & 10.4 & 10.4 & 12.3 & 12.4 & 1.4 & 6.8 & 3.4 & 9.6 & 13.3 & 12.0 \\
Search-o1 & 12.4 & 17.6 & 17.2 & 17.0 & 19.6 & 18.8 & 3.4 & 8.8 & 6.4 & 14.4 & 23.0 & 18.4 \\

\hline 
SFT & 15.8 & 20.1 & 18.2 & 28.4 & 31.1 & 30.0 & 2.2 & 9.6 & 3.8 & 8.0 & 15.8 & 8.8 \\
SimpleDeepSearcher & 34.4 & 45.3 & 47.6 & 39.6 & 46.5 & 47.8 & 12.4 & 20.5 & 18.8 & 33.6 & 44.1 & 42.4 \\
\hline
ReSearch-Base & 28.8 & 38.4 & 36.2 & 37.2 & 40.7 & 40.0 & 14.4 & 24.2 & 17.4 & 34.8 & 45.5 & 37.2 \\
ReSearch-Instruct & 30.8 & 41.7 & 43.6 & 38.0 & 42.0 & 41.6 & 14.2 & 23.8 & 18.0 & 34.8 & 47.1 & 42.0 \\
R1-search-Base & 30.8 & 40.7 & 48.0 & 37.2 & 41.1 & 40.2 & 13.6 & 23.9 & 17.0 & 33.2 & 41.3 & 37.2 \\
R1-search-Instruct & 31.2 & 42.2 & 43.4 & 42.6 & 47.1 & 46.2 & 15.4 & 25.3 & 18.6 & 33.2 & 43.5 & 39.6 \\
Search-R1-Base & 35.4 & 50.1 & 51.2 & 44.0 & 50.9 & 51.4 & 14.8 & 26.0 & 22.6 & 38.4 & 50.9 & 48.4 \\
Search-R1-Instruct & 37.4 & 49.3 & 50.4 & 47.6 & 51.7 & 53.4 & 16.2 & 23.7 & 21.0 & 40.2 & 50.3 & 47.4 \\
LeTS-Instruct* &37.1& - &55.2 &41.0& - &47.5&17.5 &-&26.9&38.4&-&51.2 \\
\hline 
\rowcolor{yellow!20}
\textbf{TIRESRAG-R1-Base} & \textbf{41.0} & 53.0 &  \textbf{56.4} & 52.4 & 58.4 & 60.2 & 16.2 & 26.9 & 24.4 & 40.4 & 52.6 & 50.0 \\
\rowcolor{yellow!20}
\textbf{TIRESRAG-R1-Instruct} & 41.0 &  \textbf{54.2} & 56.0 &  \textbf{52.8} &  \textbf{59.6} &  \textbf{61.4} &  \textbf{19.4} &  \textbf{30.0} &  \textbf{27.4} &  \textbf{44.0} &  \textbf{54.7} &  \textbf{52.8} \\
\bottomrule
\end{tabular}
\caption{\vic{Main experimental results on HotpotQA, 2WikiMultiHopQA, Musique, and Bamboogle.}}
\label{tab:main}
\end{table*}

\section{Experiments}
\label{sec:exp}
\noindent \textbf{Experimental Setup.}
\vic{Following the evaluation protocol of R1-Searcher, we assess performance on four widely used and challenging multi-hop QA datasets: \textbf{HotpotQA} \cite{yang-etal-2018-hotpotqa}, \textbf{2WikiMultiHopQA} \cite{ho-etal-2020-constructing}, \textbf{Bamboogle} \cite{press-etal-2023-measuring}, and \textbf{Musique} \cite{trivedi-etal-2022-musique}.  
For HotpotQA, 2WikiMultiHopQA, and Musique, we adopt the test splits released by R1-Searcher, each containing 500 examples. For Bamboogle, we use the full set of 125 test examples.  For training, we use the dataset provided by R1-Searcher, which includes 4,561 examples from the HotpotQA training set and 3,587 examples from the 2WikiMultiHopQA training set.}

\noindent \textbf{Evaluation Metrics.}  
\vic{We evaluate model performance using four metrics: \textbf{Exact Match (EM), F1, LLM-as-Judge, and Cover Exact Match (CEM)}. EM checks exact matches, F1 accounts for partial overlaps, LLM-as-Judge (via GPT‑4o) assesses semantic correctness, and CEM measures whether the gold answer is covered. For more detailed settings, please refer to App.~\ref{eva_metrcis}.}


\noindent \textbf{Baselines. }
\vic{We compare \textbf{TIRESRAG-R1} with 14 representative baselines spanning four categories: 
(1) \textbf{Naive prompt methods:} Direct, COT, and R1-based; 
(2) \textbf{Retrieval-augmented prompt methods:} Naive RAG, Agentic-R1, Search-o1, SURE, IRCOT, Self-Ask, and RQRAG; 
(3) \textbf{SFT methods:} SFT and SimpleDeepSearcher; 
(4) \textbf{RL methods:} Search-R1, R1-Searcher, Research, and the process-reward method LeTS. 
For detailed descriptions and configurations of these baselines, please refer to the App.~\ref{sec:app_baseline}.}

\subsection{Main Results}
 \vic{Table~\ref{tab:main} shows that \textbf{TIRESRAG-R1} achieves superior or competitive performance on all four complex multi-hop datasets when trained on either Qwen‑2.5‑3B‑Base or Qwen‑2.5‑3B‑Instruct. } 
\vic{Our key findings are as follows:}

\noindent \vic{\textbf{(1)} For the small 3B models, prompt-based methods generally perform poorly. For instance, \textit{Search-o1} even underperforms \textit{naive RAG}, as 3B models struggle to interpret instructions and generate effective retrieval queries without fine-tuning.} 

\noindent \textbf{(2)} \vic{Incorporating format rewards may not be optimal. Compared to Search-R1 (answer reward only), {Research} and {R1-Searcher} exhibit average EM drops of 5.13\% and 4.60\%, respectively. We attribute this decline to format-based learning, which reduces the exploration capacity of the 3B models and weakens useful reward signals when answers are correct but deviate slightly in format.}

\noindent \textbf{(3)} \vic{TIRESRAG-R1 delivers substantial performance gains. On in-domain datasets \textsc{HotpotQA} and \textsc{2WikiMultiHopQA}, it outperforms \textsc{Search-R1} by average EM margins of 4.7\% and 7.0\%, respectively. For out-of-domain datasets \textsc{Musique} and \textsc{Bamboogle}, we observe improvements of 5.3\% and 4.6\%.
Compared to LeTS, TIRESRAG-R1 achieves additional gains of 5.8 and 4.2 points in EM and \textsc{LLM-as-Judge}, respectively.
These results indicate that our approach effectively learns higher-quality reasoning chains and generalizes well to unseen domains.}

\begin{table*}
\centering
\scriptsize
    \begin{tabular}{l|cccccccccccc}
\toprule
Method & \multicolumn{3}{c}{Hotpotqa} & \multicolumn{3}{c}{2wikimultihopqa} & \multicolumn{3}{c}{Musique} & \multicolumn{3}{c}{Bamboogle} \\
 & EM & F1 & CEM & EM & F1 & CEM & EM & F1 & CEM & EM & F1 & CEM \\
\midrule
Ours-Base+GRPO & 41.0 & 53.0 & 47.8 & 52.4 & 58.4 & 59.0 & 16.2 & 26.9 & 21.4 & 40.4 & 52.6 & 43.6 \\

Ours-Instruct+GRPO & 41.0 & 54.2 & 46.0 & \textbf{52.8} & \textbf{59.6} & \textbf{60.8} & \textbf{19.4} & \textbf{30.0} & \textbf{23.2} & \textbf{44.0} & \textbf{54.8} & \textbf{47.2} \\
Ours-Base+Reinforce++  & \textbf{42.2} & \textbf{55.9} & 49.0 & 52.0 & 58.6 & 61.2 & 16.6 & 28.7 & 22.2 & 36.8 & 50.7 & 44.0 \\
Ours-Instruct+Reinforce++ & 39.6 & 53.9 & \textbf{49.8} & 46.2 & 55.0 & 58.4 & 15.4 & 25.4 & 20.6 & 37.6 & 48.8 & 42.4 \\
\bottomrule
\end{tabular}
\caption{\vic{Comparison between GRPO and Reinforce++ across datasets.}}
\label{tab:re++}
\vspace{-0.2cm}
\end{table*}

\begin{table*}[t]
\scriptsize
\centering
\begin{tabular}{l|cccc|c}
\toprule
\textbf{Datasets} & \textbf{HotpotQA} & \textbf{2WikiMultiHopQA} & \textbf{MusiQue} & \textbf{Bamboogle} & \textbf{Average} \\
\midrule
\multicolumn{6}{l}{\textbf{Thinking Length}} \\
Naive grpo & 318.7 & 293.1 & 467.7 & 215.2 & 323.7 \\
Search-R1-Instruct     & 258.4 & 331.3 & 360.8 & 194.3 & 286.2 \\
\rowcolor{yellow!20}
\textbf{TIRESRAG-R1-Instruct}                 & \textbf{252.1} & \textbf{312.1} & \textbf{303.6} & \textbf{229.2} & \textbf{274.3} \\

\midrule
\multicolumn{6}{l}{\textbf{Search Steps}} \\
Naive grpo & 2.7 & 2.9 & 3.9 & 2.2 & 2.93 \\
Search-R1-Instruct    & 2.3 & 2.3 & 2.8 & 2.2 & 2.40 \\
\rowcolor{yellow!20}
\textbf{TIRESRAG-R1-Instruct}                & \textbf{2.1} & \textbf{2.7} & \textbf{2.6} & \textbf{2.0} & \textbf{2.35} \\

\bottomrule
\end{tabular}
\caption{\vic{Comparison of average thinking length and search steps across datasets.}}
\label{tab:thinking_search_eff}
\end{table*}


\begin{table}[h]
\centering
\scriptsize
\begin{tabular}{lccc}
\toprule
\textbf{Method} & \textbf{NQ} & \textbf{PopQA} & \textbf{TriviaQA} \\
\midrule
COT & 10.5 & 9.7 & 7.8 \\
Sure & 25.5 & 30.4 & 13.4 \\
SimpleDeepSearcher & 33.4 & 38.9 & 59.6 \\
ReSearch-Instruct & 35.8 & 41.8 & 58.4 \\
\rowcolor{yellow!20}
\textbf{TIRESRAG-R1-Instruct} & \textbf{38.0} & \textbf{43.0} & \textbf{60.0} \\
\bottomrule
\end{tabular}
\caption{\vic{Generalization results on single-hop benchmarks.}}
\label{tab:simple_single_hop}
\vspace{-0.4cm}
\end{table}


\subsection{Analysis}
\vic{We perform a comprehensive analysis to better understand the factors influencing our method’s effectiveness. Below, we highlight the most significant findings; for a more detailed discussion and a case study, please refer to Appendix~\ref{sec:analy}.
}

\smallskip\noindent \textbf{Different RL Methods. }
\vic{To assess the generality of our training strategy, we also apply it to \textbf{Reinforce++}~\cite{hu2025reinforceefficientrlhfalgorithm} that does not use a critic model. The key difference between {Reinforce++} and GRPO is that it normalizes advantages across the entire batch rather than within group rollouts.}
\vic{We train both Qwen2.5-3B-Base and Instruct models under identical settings. As shown in Table~\ref{tab:re++}, compared with GRPO’s group-normalization strategy, Reinforce++ performs better on in-domain datasets (HotpotQA, 2WikiMultiHopQA) but worse on out-of-domain datasets (Musique, Bamboogle), which is consistent with the findings in R1-Searcher.}

\smallskip \noindent \textbf{Efficiency Analysis. }
\vic{We compare search counts and thought lengths with naive GRPO and Search-R1. As shown in Table~\ref{tab:thinking_search_eff}, despite encouraging reflection and sufficient information gathering, our method does not hurt efficiency.  
Compared to naive GRPO, our method reduces average search count by 0.58 points and token generation by 49.4 points.  
Compared to Search-R1, search count decreases by 0.05 points and token generation by 12 points.  
This indicates that RL-trained Agent-RAG models can achieve improved reasoning quality without sacrificing efficiency.}
\subsection{Generalization on Single-Hop Benchmarks}
\vic{Since our primary training data consist of multi-hop QA examples, we also evaluate on single-hop tasks to assess generalization. We use three widely adopted single-hop QA benchmarks: {NQ}, {PopQA}, and {TriviaQA} and present comparisons with the best baseline in each category.  More comprehensive results can be found in the App.~\ref{sec:app_gen_single}.
Table~\ref{tab:simple_single_hop} shows that our method consistently outperforms all baselines across all metrics. 
On \textsc{NQ}, it achieves EM score of 38.0, exceeding the strongest baseline ({ReSearch-Instruct}) by 2.2 points.
On {PopQA}, we obtain 43.0, surpassing baselines by 1.5 points.
For {TriviaQA}, we achieve 60.0, with an improvement of 2.1 points.
These results demonstrate that, despite being trained primarily on multi-hop data, our method generalizes effectively to single-hop tasks, highlighting its robustness across reasoning types and task distribution. }

\section{Conclusion}
\vic{In this work, we reveal the limitations of current outcome-supervised RL-trained RAG reasoning models.  
To address these issues, we propose \textbf{TIRESRAG-R1}, which uses a novel \textit{think–search–reflect} paradigm, explicitly encouraging models to reflect on uncertain answers.  
We design multiple reward functions to improve document retrieval sufficiency and reasoning quality, and introduce a difficulty-aware advantage strategy to provide stronger learning signals on hard problems.  
Comprehensive evaluations show that TIRESRAG-R1 outperforms existing methods on four multi-hop QA benchmarks.
Further analysis highlights that our method is compatible with different RL algorithms and exhibits strong generalization. } 
\section*{Limitations}
\textbf{Model scaling.} \vic{Our experiments are conducted only on Qwen2.5-3B due to computational constraints. Although TIRESRAG-R1 shows promising results on smaller models, its effectiveness on larger architectures (e.g., Qwen2.5-7B) remains unexplored.}

\smallskip \noindent\textbf{Reward modeling.} \vic{We use Qwen3-8B for sufficient and thinking scoring. Although it provides accurate signals, using a stronger model such as GPT-4o, or a specialized reward model fine-tuned on domain-specific data, may further improve performance. Exploring more accurate reward models (e.g., as suggested in recent thinking-reward literature) is a promising direction.}

\smallskip \noindent\textbf{Reflection signal sparsity.} \vic{While our reflection mechanism corrects some wrong answers, the number of training examples requiring reflection is limited, reducing useful learning signals. Future work could synthesize reflection-rich data for SFT before applying TIRESRAG-R1 to further improve the model’s ability to reflect effectively.}

\bibliography{acl_latex}

\begin{thebibliography}{65}
\providecommand{\natexlab}[1]{#1}

\bibitem[{Ahn et~al.(2024)Ahn, Verma, Lou, Liu, Zhang, and Yin}]{ahn-etal-2024-large}
Janice Ahn, Rishu Verma, Renze Lou, Di~Liu, Rui Zhang, and Wenpeng Yin. 2024.
\newblock \href {https://aclanthology.org/2024.eacl-srw.17/} {Large language models for mathematical reasoning: Progresses and challenges}.
\newblock In \emph{Proceedings of the 18th Conference of the European Chapter of the Association for Computational Linguistics: Student Research Workshop}, pages 225--237, St. Julian{'}s, Malta. Association for Computational Linguistics.

\bibitem[{Ayala and Bechard(2024)}]{ayala-bechard-2024-reducing}
Orlando Ayala and Patrice Bechard. 2024.
\newblock \href {https://doi.org/10.18653/v1/2024.naacl-industry.19} {Reducing hallucination in structured outputs via retrieval-augmented generation}.
\newblock In \emph{Proceedings of the 2024 Conference of the North American Chapter of the Association for Computational Linguistics: Human Language Technologies (Volume 6: Industry Track)}, pages 228--238, Mexico City, Mexico. Association for Computational Linguistics.

\bibitem[{Chan et~al.(2024)Chan, Xu, Yuan, Luo, Xue, Guo, and Fu}]{chan2024rqrag}
Chi-Min Chan, Chunpu Xu, Ruibin Yuan, Hongyin Luo, Wei Xue, Yike Guo, and Jie Fu. 2024.
\newblock \href {https://openreview.net/forum?id=tzE7VqsaJ4} {{RQ}-{RAG}: Learning to refine queries for retrieval augmented generation}.
\newblock In \emph{First Conference on Language Modeling}.

\bibitem[{Chen et~al.(2025)Chen, Li, Sun, Zhou, Zhu, Wang, Pan, Zhang, Chen, Yang, Zhou, and Chen}]{chen2025researchlearningreasonsearch}
Mingyang Chen, Tianpeng Li, Haoze Sun, Yijie Zhou, Chenzheng Zhu, Haofen Wang, Jeff~Z. Pan, Wen Zhang, Huajun Chen, Fan Yang, Zenan Zhou, and Weipeng Chen. 2025.
\newblock \href {https://arxiv.org/abs/2503.19470} {Research: Learning to reason with search for llms via reinforcement learning}.
\newblock \emph{Preprint}, arXiv:2503.19470.

\bibitem[{DeepSeek-AI et~al.(2025)DeepSeek-AI, Guo, Yang, Zhang, Song, Zhang, Xu, Zhu, Ma, Wang, Bi, Zhang, Yu, Wu, Wu, Gou, Shao, Li, Gao, and Aixin Liu~etc}]{deepseekai2025deepseekr1incentivizingreasoningcapability}
DeepSeek-AI, Daya Guo, Dejian Yang, Haowei Zhang, Junxiao Song, Ruoyu Zhang, Runxin Xu, Qihao Zhu, Shirong Ma, Peiyi Wang, Xiao Bi, Xiaokang Zhang, Xingkai Yu, Yu~Wu, Z.~F. Wu, Zhibin Gou, Zhihong Shao, Zhuoshu Li, Ziyi Gao, and .~Aixin Liu~etc. 2025.
\newblock \href {https://arxiv.org/abs/2501.12948} {Deepseek-r1: Incentivizing reasoning capability in llms via reinforcement learning}.
\newblock \emph{Preprint}, arXiv:2501.12948.

\bibitem[{Gade and Jetcheva(2024)}]{gade2024itstimeincorporatingtemporality}
Anoushka Gade and Jorjeta Jetcheva. 2024.
\newblock \href {https://arxiv.org/abs/2401.13222} {It's about time: Incorporating temporality in retrieval augmented language models}.
\newblock \emph{Preprint}, arXiv:2401.13222.

\bibitem[{Gao et~al.(2024{\natexlab{a}})Gao, Xiong, Gao, Jia, Pan, Bi, Dai, Sun, Wang, and Wang}]{gao_retrieval-augmented_2024}
Yunfan Gao, Yun Xiong, Xinyu Gao, Kangxiang Jia, Jinliu Pan, Yuxi Bi, Yi~Dai, Jiawei Sun, Meng Wang, and Haofen Wang. 2024{\natexlab{a}}.
\newblock \href {https://doi.org/10.48550/arXiv.2312.10997} {Retrieval-{Augmented} {Generation} for {Large} {Language} {Models}: {A} {Survey}}.
\newblock \emph{arXiv preprint}.
\newblock ArXiv:2312.10997.

\bibitem[{Gao et~al.(2024{\natexlab{b}})Gao, Xiong, Wang, and Wang}]{gao2024modularragtransformingrag}
Yunfan Gao, Yun Xiong, Meng Wang, and Haofen Wang. 2024{\natexlab{b}}.
\newblock \href {https://arxiv.org/abs/2407.21059} {Modular rag: Transforming rag systems into lego-like reconfigurable frameworks}.
\newblock \emph{Preprint}, arXiv:2407.21059.

\bibitem[{Grattafiori et~al.(2024)Grattafiori, Dubey, Jauhri, Pandey, Kadian, Al-Dahle, Letman, Mathur, Schelten, Vaughan, Yang, Fan, Goyal, Hartshorn, Yang, Mitra, Sravankumar, Korenev, Hinsvark, Rao, Zhang, Rodriguez, Gregerson, Spataru, Roziere, Biron, and Binh Tang~etc}]{grattafiori2024llama3herdmodels}
Aaron Grattafiori, Abhimanyu Dubey, Abhinav Jauhri, Abhinav Pandey, Abhishek Kadian, Ahmad Al-Dahle, Aiesha Letman, Akhil Mathur, Alan Schelten, Alex Vaughan, Amy Yang, Angela Fan, Anirudh Goyal, Anthony Hartshorn, Aobo Yang, Archi Mitra, Archie Sravankumar, Artem Korenev, Arthur Hinsvark, Arun Rao, Aston Zhang, Aurelien Rodriguez, Austen Gregerson, Ava Spataru, Baptiste Roziere, Bethany Biron, and .~Binh Tang~etc. 2024.
\newblock \href {https://arxiv.org/abs/2407.21783} {The llama 3 herd of models}.
\newblock \emph{Preprint}, arXiv:2407.21783.

\bibitem[{Guu et~al.(2020)Guu, Lee, Tung, Pasupat, and Chang}]{10.5555/3524938.3525306}
Kelvin Guu, Kenton Lee, Zora Tung, Panupong Pasupat, and Ming-Wei Chang. 2020.
\newblock Realm: retrieval-augmented language model pre-training.
\newblock In \emph{Proceedings of the 37th International Conference on Machine Learning}, ICML'20. JMLR.org.

\bibitem[{Ho et~al.(2020)Ho, Duong~Nguyen, Sugawara, and Aizawa}]{ho-etal-2020-constructing}
Xanh Ho, Anh-Khoa Duong~Nguyen, Saku Sugawara, and Akiko Aizawa. 2020.
\newblock \href {https://doi.org/10.18653/v1/2020.coling-main.580} {Constructing a multi-hop {QA} dataset for comprehensive evaluation of reasoning steps}.
\newblock In \emph{Proceedings of the 28th International Conference on Computational Linguistics}, pages 6609--6625, Barcelona, Spain (Online). International Committee on Computational Linguistics.

\bibitem[{Hu et~al.(2025{\natexlab{a}})Hu, Liu, and Shen}]{hu2025reinforceefficientrlhfalgorithm}
Jian Hu, Jason~Klein Liu, and Wei Shen. 2025{\natexlab{a}}.
\newblock \href {https://arxiv.org/abs/2501.03262} {Reinforce++: An efficient rlhf algorithm with robustness to both prompt and reward models}.
\newblock \emph{Preprint}, arXiv:2501.03262.

\bibitem[{Hu et~al.(2025{\natexlab{b}})Hu, Wu, Shen, Liu, Zhu, Wang, Jiang, Wang, Chen, Chen, Fang, Xianyu, Cao, Xu, and Liu}]{hu2025openrlhfeasytousescalablehighperformance}
Jian Hu, Xibin Wu, Wei Shen, Jason~Klein Liu, Zilin Zhu, Weixun Wang, Songlin Jiang, Haoran Wang, Hao Chen, Bin Chen, Weikai Fang, Xianyu, Yu~Cao, Haotian Xu, and Yiming Liu. 2025{\natexlab{b}}.
\newblock \href {https://arxiv.org/abs/2405.11143} {Openrlhf: An easy-to-use, scalable and high-performance rlhf framework}.
\newblock \emph{Preprint}, arXiv:2405.11143.

\bibitem[{Hu et~al.(2025{\natexlab{c}})Hu, Zhang, Han, Jiang, Zhang, and Shum}]{hu2025openreasonerzeroopensourceapproach}
Jingcheng Hu, Yinmin Zhang, Qi~Han, Daxin Jiang, Xiangyu Zhang, and Heung-Yeung Shum. 2025{\natexlab{c}}.
\newblock \href {https://arxiv.org/abs/2503.24290} {Open-reasoner-zero: An open source approach to scaling up reinforcement learning on the base model}.
\newblock \emph{Preprint}, arXiv:2503.24290.

\bibitem[{Jeong et~al.(2024)Jeong, Baek, Cho, Hwang, and Park}]{jeong-etal-2024-adaptive}
Soyeong Jeong, Jinheon Baek, Sukmin Cho, Sung~Ju Hwang, and Jong Park. 2024.
\newblock \href {https://doi.org/10.18653/v1/2024.naacl-long.389} {Adaptive-{RAG}: Learning to adapt retrieval-augmented large language models through question complexity}.
\newblock In \emph{Proceedings of the 2024 Conference of the North American Chapter of the Association for Computational Linguistics: Human Language Technologies (Volume 1: Long Papers)}, pages 7036--7050, Mexico City, Mexico. Association for Computational Linguistics.

\bibitem[{Jiang et~al.(2025)Jiang, Wang, Shen, Kim, and Kim}]{10.1145/3747588}
Juyong Jiang, Fan Wang, Jiasi Shen, Sungju Kim, and Sunghun Kim. 2025.
\newblock \href {https://doi.org/10.1145/3747588} {A survey on large language models for code generation}.
\newblock \emph{ACM Trans. Softw. Eng. Methodol.}
\newblock Just Accepted.

\bibitem[{Jin et~al.(2025{\natexlab{a}})Jin, Zeng, Yue, Yoon, Arik, Wang, Zamani, and Han}]{jin2025searchr1trainingllmsreason}
Bowen Jin, Hansi Zeng, Zhenrui Yue, Jinsung Yoon, Sercan Arik, Dong Wang, Hamed Zamani, and Jiawei Han. 2025{\natexlab{a}}.
\newblock \href {https://arxiv.org/abs/2503.09516} {Search-r1: Training llms to reason and leverage search engines with reinforcement learning}.
\newblock \emph{Preprint}, arXiv:2503.09516.

\bibitem[{Jin et~al.(2025{\natexlab{b}})Jin, Zhu, Dou, Dong, Yang, Zhang, Zhao, Yang, and Wen}]{10.1145/3701716.3715313}
Jiajie Jin, Yutao Zhu, Zhicheng Dou, Guanting Dong, Xinyu Yang, Chenghao Zhang, Tong Zhao, Zhao Yang, and Ji-Rong Wen. 2025{\natexlab{b}}.
\newblock \href {https://doi.org/10.1145/3701716.3715313} {Flashrag: A modular toolkit for efficient retrieval-augmented generation research}.
\newblock In \emph{Companion Proceedings of the ACM on Web Conference 2025}, WWW '25, page 737–740, New York, NY, USA. Association for Computing Machinery.

\bibitem[{Joren et~al.(2025)Joren, Zhang, Ferng, Juan, Taly, and Rashtchian}]{joren2025sufficient}
Hailey Joren, Jianyi Zhang, Chun-Sung Ferng, Da-Cheng Juan, Ankur Taly, and Cyrus Rashtchian. 2025.
\newblock \href {https://openreview.net/forum?id=Jjr2Odj8DJ} {Sufficient context: A new lens on retrieval augmented generation systems}.
\newblock In \emph{The Thirteenth International Conference on Learning Representations}.

\bibitem[{Kamalloo et~al.(2023)Kamalloo, Dziri, Clarke, and Rafiei}]{kamalloo-etal-2023-evaluating}
Ehsan Kamalloo, Nouha Dziri, Charles Clarke, and Davood Rafiei. 2023.
\newblock \href {https://doi.org/10.18653/v1/2023.acl-long.307} {Evaluating open-domain question answering in the era of large language models}.
\newblock In \emph{Proceedings of the 61st Annual Meeting of the Association for Computational Linguistics (Volume 1: Long Papers)}, pages 5591--5606, Toronto, Canada. Association for Computational Linguistics.

\bibitem[{Kim et~al.(2024)Kim, Nam, Mo, Park, Lee, Seo, Ha, and Shin}]{kim2024sure}
Jaehyung Kim, Jaehyun Nam, Sangwoo Mo, Jongjin Park, Sang-Woo Lee, Minjoon Seo, Jung-Woo Ha, and Jinwoo Shin. 2024.
\newblock \href {https://openreview.net/forum?id=w4DW6qkRmt} {Sure: Summarizing retrievals using answer candidates for open-domain {QA} of {LLM}s}.
\newblock In \emph{The Twelfth International Conference on Learning Representations}.

\bibitem[{Kwon et~al.(2023)Kwon, Li, Zhuang, Sheng, Zheng, Yu, Gonzalez, Zhang, and Stoica}]{10.1145/3600006.3613165}
Woosuk Kwon, Zhuohan Li, Siyuan Zhuang, Ying Sheng, Lianmin Zheng, Cody~Hao Yu, Joseph Gonzalez, Hao Zhang, and Ion Stoica. 2023.
\newblock \href {https://doi.org/10.1145/3600006.3613165} {Efficient memory management for large language model serving with pagedattention}.
\newblock In \emph{Proceedings of the 29th Symposium on Operating Systems Principles}, SOSP '23, page 611–626, New York, NY, USA. Association for Computing Machinery.

\bibitem[{Lee et~al.(2024)Lee, An, and Kim}]{lee-etal-2024-planrag}
Myeonghwa Lee, Seonho An, and Min-Soo Kim. 2024.
\newblock \href {https://doi.org/10.18653/v1/2024.naacl-long.364} {{P}lan{RAG}: A plan-then-retrieval augmented generation for generative large language models as decision makers}.
\newblock In \emph{Proceedings of the 2024 Conference of the North American Chapter of the Association for Computational Linguistics: Human Language Technologies (Volume 1: Long Papers)}, pages 6537--6555, Mexico City, Mexico. Association for Computational Linguistics.

\bibitem[{Lewis et~al.(2020)Lewis, Perez, Piktus, Petroni, Karpukhin, Goyal, K\"{u}ttler, Lewis, Yih, Rockt\"{a}schel, Riedel, and Kiela}]{NEURIPS2020_6b493230}
Patrick Lewis, Ethan Perez, Aleksandra Piktus, Fabio Petroni, Vladimir Karpukhin, Naman Goyal, Heinrich K\"{u}ttler, Mike Lewis, Wen-tau Yih, Tim Rockt\"{a}schel, Sebastian Riedel, and Douwe Kiela. 2020.
\newblock \href {https://proceedings.neurips.cc/paper_files/paper/2020/file/6b493230205f780e1bc26945df7481e5-Paper.pdf} {Retrieval-augmented generation for knowledge-intensive nlp tasks}.
\newblock In \emph{Advances in Neural Information Processing Systems}, volume~33, pages 9459--9474. Curran Associates, Inc.

\bibitem[{Li et~al.(2024{\natexlab{a}})Li, Yuan, and Zhang}]{li2024enhancingllmfactualaccuracy}
Jiarui Li, Ye~Yuan, and Zehua Zhang. 2024{\natexlab{a}}.
\newblock \href {https://arxiv.org/abs/2403.10446} {Enhancing llm factual accuracy with rag to counter hallucinations: A case study on domain-specific queries in private knowledge-bases}.
\newblock \emph{Preprint}, arXiv:2403.10446.

\bibitem[{Li et~al.(2025{\natexlab{a}})Li, Dong, Jin, Zhang, Zhou, Zhu, Zhang, and Dou}]{li2025searcho1agenticsearchenhancedlarge}
Xiaoxi Li, Guanting Dong, Jiajie Jin, Yuyao Zhang, Yujia Zhou, Yutao Zhu, Peitian Zhang, and Zhicheng Dou. 2025{\natexlab{a}}.
\newblock \href {https://arxiv.org/abs/2501.05366} {Search-o1: Agentic search-enhanced large reasoning models}.
\newblock \emph{Preprint}, arXiv:2501.05366.

\bibitem[{Li et~al.(2024{\natexlab{b}})Li, Xu, Zhao, Jiao, Joty, and Bing}]{li2024elicitreasoningllmscriticguided}
Xingxuan Li, Weiwen Xu, Ruochen Zhao, Fangkai Jiao, Shafiq Joty, and Lidong Bing. 2024{\natexlab{b}}.
\newblock \href {https://arxiv.org/abs/2410.01428} {Can we further elicit reasoning in llms? critic-guided planning with retrieval-augmentation for solving challenging tasks}.
\newblock \emph{Preprint}, arXiv:2410.01428.

\bibitem[{Li et~al.(2025{\natexlab{b}})Li, Luo, Li, Li, Cheng, Wang, Zheng, Wang, Yin, and Qiu}]{li2025r3raglearningstepbystepreasoning}
Yuan Li, Qi~Luo, Xiaonan Li, Bufan Li, Qinyuan Cheng, Bo~Wang, Yining Zheng, Yuxin Wang, Zhangyue Yin, and Xipeng Qiu. 2025{\natexlab{b}}.
\newblock \href {https://arxiv.org/abs/2505.23794} {R3-rag: Learning step-by-step reasoning and retrieval for llms via reinforcement learning}.
\newblock \emph{Preprint}, arXiv:2505.23794.

\bibitem[{Liu et~al.(2025)Liu, Chen, Li, Qi, Pang, Du, Lee, and Lin}]{liu2025understandingr1zeroliketrainingcritical}
Zichen Liu, Changyu Chen, Wenjun Li, Penghui Qi, Tianyu Pang, Chao Du, Wee~Sun Lee, and Min Lin. 2025.
\newblock \href {https://arxiv.org/abs/2503.20783} {Understanding r1-zero-like training: A critical perspective}.
\newblock \emph{Preprint}, arXiv:2503.20783.

\bibitem[{Meng et~al.(2024)Meng, Song, Tong, Pan, and Yu}]{10.1109/ASE56229.2023.00038}
Chunyang Meng, Shijie Song, Haogang Tong, Maolin Pan, and Yang Yu. 2024.
\newblock \href {https://doi.org/10.1109/ASE56229.2023.00038} {Deepscaler: Holistic autoscaling for microservices based on spatiotemporal gnn with adaptive graph learning}.
\newblock In \emph{Proceedings of the 38th IEEE/ACM International Conference on Automated Software Engineering}, ASE '23, page 53–65. IEEE Press.

\bibitem[{Mousavi et~al.(2024)Mousavi, Alghisi, and Riccardi}]{mousavi2024dyknowdynamicallyverifyingtimesensitive}
Seyed~Mahed Mousavi, Simone Alghisi, and Giuseppe Riccardi. 2024.
\newblock \href {https://arxiv.org/abs/2404.08700} {Dyknow: Dynamically verifying time-sensitive factual knowledge in llms}.
\newblock \emph{Preprint}, arXiv:2404.08700.

\bibitem[{OpenAI et~al.(2024{\natexlab{a}})OpenAI, :, Hurst, Lerer, Goucher, Perelman, Ramesh, Clark, and AJ~Ostrow~etc}]{openai2024gpt4ocard}
OpenAI, :, Aaron Hurst, Adam Lerer, Adam~P. Goucher, Adam Perelman, Aditya Ramesh, Aidan Clark, and .~AJ~Ostrow~etc. 2024{\natexlab{a}}.
\newblock \href {https://arxiv.org/abs/2410.21276} {Gpt-4o system card}.
\newblock \emph{Preprint}, arXiv:2410.21276.

\bibitem[{OpenAI et~al.(2024{\natexlab{b}})OpenAI, :, Jaech, Kalai, Lerer, Richardson, El-Kishky, Low, Helyar, Madry, Beutel, Carney, Iftimie, and Alex Karpenko~etc}]{openai2024openaio1card}
OpenAI, :, Aaron Jaech, Adam Kalai, Adam Lerer, Adam Richardson, Ahmed El-Kishky, Aiden Low, Alec Helyar, Aleksander Madry, Alex Beutel, Alex Carney, Alex Iftimie, and .~Alex Karpenko~etc. 2024{\natexlab{b}}.
\newblock \href {https://arxiv.org/abs/2412.16720} {Openai o1 system card}.
\newblock \emph{Preprint}, arXiv:2412.16720.

\bibitem[{Pan et~al.(2023)Pan, Albalak, Wang, and Wang}]{pan-etal-2023-logic}
Liangming Pan, Alon Albalak, Xinyi Wang, and William Wang. 2023.
\newblock \href {https://doi.org/10.18653/v1/2023.findings-emnlp.248} {Logic-{LM}: Empowering large language models with symbolic solvers for faithful logical reasoning}.
\newblock In \emph{Findings of the Association for Computational Linguistics: EMNLP 2023}, pages 3806--3824, Singapore. Association for Computational Linguistics.

\bibitem[{Petroni et~al.(2021)Petroni, Piktus, Fan, Lewis, Yazdani, De~Cao, Thorne, Jernite, Karpukhin, Maillard, Plachouras, Rockt{\"a}schel, and Riedel}]{petroni-etal-2021-kilt}
Fabio Petroni, Aleksandra Piktus, Angela Fan, Patrick Lewis, Majid Yazdani, Nicola De~Cao, James Thorne, Yacine Jernite, Vladimir Karpukhin, Jean Maillard, Vassilis Plachouras, Tim Rockt{\"a}schel, and Sebastian Riedel. 2021.
\newblock \href {https://doi.org/10.18653/v1/2021.naacl-main.200} {{KILT}: a benchmark for knowledge intensive language tasks}.
\newblock In \emph{Proceedings of the 2021 Conference of the North American Chapter of the Association for Computational Linguistics: Human Language Technologies}, pages 2523--2544, Online. Association for Computational Linguistics.

\bibitem[{Press et~al.(2023)Press, Zhang, Min, Schmidt, Smith, and Lewis}]{press-etal-2023-measuring}
Ofir Press, Muru Zhang, Sewon Min, Ludwig Schmidt, Noah Smith, and Mike Lewis. 2023.
\newblock \href {https://doi.org/10.18653/v1/2023.findings-emnlp.378} {Measuring and narrowing the compositionality gap in language models}.
\newblock In \emph{Findings of the Association for Computational Linguistics: EMNLP 2023}, pages 5687--5711, Singapore. Association for Computational Linguistics.

\bibitem[{Ren et~al.(2025)Ren, Xu, Wang, Li, Ma, and Liu}]{ren2025effectivetransparentragadaptivereward}
Jingyi Ren, Yekun Xu, Xiaolong Wang, Weitao Li, Weizhi Ma, and Yang Liu. 2025.
\newblock \href {https://arxiv.org/abs/2505.13258} {Effective and transparent rag: Adaptive-reward reinforcement learning for decision traceability}.
\newblock \emph{Preprint}, arXiv:2505.13258.

\bibitem[{Sha et~al.(2025)Sha, Cui, and Wang}]{sha2025semreinforcementlearningsearchefficient}
Zeyang Sha, Shiwen Cui, and Weiqiang Wang. 2025.
\newblock \href {https://arxiv.org/abs/2505.07903} {Sem: Reinforcement learning for search-efficient large language models}.
\newblock \emph{Preprint}, arXiv:2505.07903.

\bibitem[{Shao et~al.(2023)Shao, Gong, Shen, Huang, Duan, and Chen}]{shao-etal-2023-enhancing}
Zhihong Shao, Yeyun Gong, Yelong Shen, Minlie Huang, Nan Duan, and Weizhu Chen. 2023.
\newblock \href {https://doi.org/10.18653/v1/2023.findings-emnlp.620} {Enhancing retrieval-augmented large language models with iterative retrieval-generation synergy}.
\newblock In \emph{Findings of the Association for Computational Linguistics: EMNLP 2023}, pages 9248--9274, Singapore. Association for Computational Linguistics.

\bibitem[{Shao et~al.(2024)Shao, Wang, Qihao~Zhu, Song, Zhang, Li, Wu, and Guo}]{deepseek-math}
Zhihong Shao, Peiyi Wang, Runxin~Xu Qihao~Zhu, Junxiao Song, Mingchuan Zhang, Y.K. Li, Y.~Wu, and Daya Guo. 2024.
\newblock \href {https://arxiv.org/abs/2402.03300} {Deepseekmath: Pushing the limits of mathematical reasoning in open language models}.

\bibitem[{Shi et~al.(2025)Shi, Li, Wu, Liu, Fang, Cai, Zhang, and Wang}]{shi2025searchrefinethinkautonomous}
Yaorui Shi, Sihang Li, Chang Wu, Zhiyuan Liu, Junfeng Fang, Hengxing Cai, An~Zhang, and Xiang Wang. 2025.
\newblock \href {https://arxiv.org/abs/2505.11277} {Search and refine during think: Autonomous retrieval-augmented reasoning of llms}.
\newblock \emph{Preprint}, arXiv:2505.11277.

\bibitem[{Siriwardhana et~al.(2023)Siriwardhana, Weerasekera, Wen, Kaluarachchi, Rana, and Nanayakkara}]{siriwardhana-etal-2023-improving}
Shamane Siriwardhana, Rivindu Weerasekera, Elliott Wen, Tharindu Kaluarachchi, Rajib Rana, and Suranga Nanayakkara. 2023.
\newblock \href {https://doi.org/10.1162/tacl_a_00530} {Improving the domain adaptation of retrieval augmented generation ({RAG}) models for open domain question answering}.
\newblock \emph{Transactions of the Association for Computational Linguistics}, 11:1--17.

\bibitem[{Snell et~al.(2025)Snell, Lee, Xu, and Kumar}]{snell2025scaling}
Charlie~Victor Snell, Jaehoon Lee, Kelvin Xu, and Aviral Kumar. 2025.
\newblock \href {https://openreview.net/forum?id=4FWAwZtd2n} {Scaling {LLM} test-time compute optimally can be more effective than scaling parameters for reasoning}.
\newblock In \emph{The Thirteenth International Conference on Learning Representations}.

\bibitem[{Song et~al.(2025)Song, Jiang, Min, Chen, Chen, Zhao, Fang, and Wen}]{song2025r1searcherincentivizingsearchcapability}
Huatong Song, Jinhao Jiang, Yingqian Min, Jie Chen, Zhipeng Chen, Wayne~Xin Zhao, Lei Fang, and Ji-Rong Wen. 2025.
\newblock \href {https://arxiv.org/abs/2503.05592} {R1-searcher: Incentivizing the search capability in llms via reinforcement learning}.
\newblock \emph{Preprint}, arXiv:2503.05592.

\bibitem[{Sun et~al.(2025{\natexlab{a}})Sun, Qiao, Guo, Fan, Hou, Jiang, Xie, Zhang, Huang, and Zhou}]{sun2025zerosearchincentivizesearchcapability}
Hao Sun, Zile Qiao, Jiayan Guo, Xuanbo Fan, Yingyan Hou, Yong Jiang, Pengjun Xie, Yan Zhang, Fei Huang, and Jingren Zhou. 2025{\natexlab{a}}.
\newblock \href {https://arxiv.org/abs/2505.04588} {Zerosearch: Incentivize the search capability of llms without searching}.
\newblock \emph{Preprint}, arXiv:2505.04588.

\bibitem[{Sun et~al.(2025{\natexlab{b}})Sun, Song, Wang, Ren, Jiang, Zhang, Bai, Deng, Zhao, Liu, Fang, Wang, and Wen}]{sun2025simpledeepsearcherdeepinformationseeking}
Shuang Sun, Huatong Song, Yuhao Wang, Ruiyang Ren, Jinhao Jiang, Junjie Zhang, Fei Bai, Jia Deng, Wayne~Xin Zhao, Zheng Liu, Lei Fang, Zhongyuan Wang, and Ji-Rong Wen. 2025{\natexlab{b}}.
\newblock \href {https://arxiv.org/abs/2505.16834} {Simpledeepsearcher: Deep information seeking via web-powered reasoning trajectory synthesis}.
\newblock \emph{Preprint}, arXiv:2505.16834.

\bibitem[{Tang and Yang(2024)}]{tang2024multihoprag}
Yixuan Tang and Yi~Yang. 2024.
\newblock \href {https://openreview.net/forum?id=t4eB3zYWBK} {Multihop-{RAG}: Benchmarking retrieval-augmented generation for multi-hop queries}.
\newblock In \emph{First Conference on Language Modeling}.

\bibitem[{Trivedi et~al.(2022)Trivedi, Balasubramanian, Khot, and Sabharwal}]{trivedi-etal-2022-musique}
Harsh Trivedi, Niranjan Balasubramanian, Tushar Khot, and Ashish Sabharwal. 2022.
\newblock \href {https://doi.org/10.1162/tacl_a_00475} {{M}u{S}i{Q}ue: Multihop questions via single-hop question composition}.
\newblock \emph{Transactions of the Association for Computational Linguistics}, 10:539--554.

\bibitem[{Trivedi et~al.(2023)Trivedi, Balasubramanian, Khot, and Sabharwal}]{trivedi-etal-2023-interleaving}
Harsh Trivedi, Niranjan Balasubramanian, Tushar Khot, and Ashish Sabharwal. 2023.
\newblock \href {https://doi.org/10.18653/v1/2023.acl-long.557} {Interleaving retrieval with chain-of-thought reasoning for knowledge-intensive multi-step questions}.
\newblock In \emph{Proceedings of the 61st Annual Meeting of the Association for Computational Linguistics (Volume 1: Long Papers)}, pages 10014--10037, Toronto, Canada. Association for Computational Linguistics.

\bibitem[{Wang et~al.(2023)Wang, Xu, Lan, Hu, Lan, Lee, and Lim}]{wang-etal-2023-plan}
Lei Wang, Wanyu Xu, Yihuai Lan, Zhiqiang Hu, Yunshi Lan, Roy Ka-Wei Lee, and Ee-Peng Lim. 2023.
\newblock \href {https://doi.org/10.18653/v1/2023.acl-long.147} {Plan-and-solve prompting: Improving zero-shot chain-of-thought reasoning by large language models}.
\newblock In \emph{Proceedings of the 61st Annual Meeting of the Association for Computational Linguistics (Volume 1: Long Papers)}, pages 2609--2634, Toronto, Canada. Association for Computational Linguistics.

\bibitem[{Wang et~al.(2024{\natexlab{a}})Wang, Wang, Guo, Chen, Zhang, Ma, and Zheng}]{wang2024rlcoderreinforcementlearningrepositorylevel}
Yanlin Wang, Yanli Wang, Daya Guo, Jiachi Chen, Ruikai Zhang, Yuchi Ma, and Zibin Zheng. 2024{\natexlab{a}}.
\newblock \href {https://arxiv.org/abs/2407.19487} {Rlcoder: Reinforcement learning for repository-level code completion}.
\newblock \emph{Preprint}, arXiv:2407.19487.

\bibitem[{Wang et~al.(2024{\natexlab{b}})Wang, Feng, Wang, Shi, Balachandran, He, and Tsvetkov}]{wang2024resolving}
Yike Wang, Shangbin Feng, Heng Wang, Weijia Shi, Vidhisha Balachandran, Tianxing He, and Yulia Tsvetkov. 2024{\natexlab{b}}.
\newblock \href {https://openreview.net/forum?id=ptvV5HGTNN} {Resolving knowledge conflicts in large language models}.
\newblock In \emph{First Conference on Language Modeling}.

\bibitem[{Wang et~al.(2025)Wang, Zheng, An, Ouyang, Cai, Wang, and Wu}]{wang2025stepsearchignitingllmssearch}
Ziliang Wang, Xuhui Zheng, Kang An, Cijun Ouyang, Jialu Cai, Yuhang Wang, and Yichao Wu. 2025.
\newblock \href {https://arxiv.org/abs/2505.15107} {Stepsearch: Igniting llms search ability via step-wise proximal policy optimization}.
\newblock \emph{Preprint}, arXiv:2505.15107.

\bibitem[{Wei et~al.(2022)Wei, Wang, Schuurmans, Bosma, Ichter, Xia, Chi, Le, and Zhou}]{10.5555/3600270.3602070}
Jason Wei, Xuezhi Wang, Dale Schuurmans, Maarten Bosma, Brian Ichter, Fei Xia, Ed~H. Chi, Quoc~V. Le, and Denny Zhou. 2022.
\newblock Chain-of-thought prompting elicits reasoning in large language models.
\newblock In \emph{Proceedings of the 36th International Conference on Neural Information Processing Systems}, NIPS '22, Red Hook, NY, USA. Curran Associates Inc.

\bibitem[{Wen et~al.(2025)Wen, Cai, Xiao, He, An, Duan, Du, Liu, Tang, Lv, Zou, Deng, Jia, and Zhang}]{wen2025lightr1curriculumsftdpo}
Liang Wen, Yunke Cai, Fenrui Xiao, Xin He, Qi~An, Zhenyu Duan, Yimin Du, Junchen Liu, Lifu Tang, Xiaowei Lv, Haosheng Zou, Yongchao Deng, Shousheng Jia, and Xiangzheng Zhang. 2025.
\newblock \href {https://arxiv.org/abs/2503.10460} {Light-r1: Curriculum sft, dpo and rl for long cot from scratch and beyond}.
\newblock \emph{Preprint}, arXiv:2503.10460.

\bibitem[{Xu et~al.(2024)Xu, Qi, Guo, Wang, Wang, Zhang, and Xu}]{xu-etal-2024-knowledge-conflicts}
Rongwu Xu, Zehan Qi, Zhijiang Guo, Cunxiang Wang, Hongru Wang, Yue Zhang, and Wei Xu. 2024.
\newblock \href {https://doi.org/10.18653/v1/2024.emnlp-main.486} {Knowledge conflicts for {LLM}s: A survey}.
\newblock In \emph{Proceedings of the 2024 Conference on Empirical Methods in Natural Language Processing}, pages 8541--8565, Miami, Florida, USA. Association for Computational Linguistics.

\bibitem[{Yang et~al.(2025)Yang, Li, Yang, Zhang, Hui, Zheng, Yu, Gao, Huang, Lv, Zheng, Liu, Zhou, Huang, Hu, Ge, Wei, Lin, Tang, Yang, Tu, Zhang, Yang, Yang, Zhou, Zhou, Lin, Dang, Bao, Yang, Yu, Deng, Li, Xue, Li, Zhang, Wang, Zhu, Men, Gao, Liu, Luo, Li, Tang, Yin, Ren, Wang, Zhang, Ren, Fan, Su, Zhang, Zhang, Wan, Liu, Wang, Cui, Zhang, Zhou, and Qiu}]{yang2025qwen3technicalreport}
An~Yang, Anfeng Li, Baosong Yang, Beichen Zhang, Binyuan Hui, Bo~Zheng, Bowen Yu, Chang Gao, Chengen Huang, Chenxu Lv, Chujie Zheng, Dayiheng Liu, Fan Zhou, Fei Huang, Feng Hu, Hao Ge, Haoran Wei, Huan Lin, Jialong Tang, Jian Yang, Jianhong Tu, Jianwei Zhang, Jianxin Yang, Jiaxi Yang, Jing Zhou, Jingren Zhou, Junyang Lin, Kai Dang, Keqin Bao, Kexin Yang, Le~Yu, Lianghao Deng, Mei Li, Mingfeng Xue, Mingze Li, Pei Zhang, Peng Wang, Qin Zhu, Rui Men, Ruize Gao, Shixuan Liu, Shuang Luo, Tianhao Li, Tianyi Tang, Wenbiao Yin, Xingzhang Ren, Xinyu Wang, Xinyu Zhang, Xuancheng Ren, Yang Fan, Yang Su, Yichang Zhang, Yinger Zhang, Yu~Wan, Yuqiong Liu, Zekun Wang, Zeyu Cui, Zhenru Zhang, Zhipeng Zhou, and Zihan Qiu. 2025.
\newblock \href {https://arxiv.org/abs/2505.09388} {Qwen3 technical report}.
\newblock \emph{Preprint}, arXiv:2505.09388.

\bibitem[{Yang et~al.(2018)Yang, Qi, Zhang, Bengio, Cohen, Salakhutdinov, and Manning}]{yang-etal-2018-hotpotqa}
Zhilin Yang, Peng Qi, Saizheng Zhang, Yoshua Bengio, William Cohen, Ruslan Salakhutdinov, and Christopher~D. Manning. 2018.
\newblock \href {https://doi.org/10.18653/v1/D18-1259} {{H}otpot{QA}: A dataset for diverse, explainable multi-hop question answering}.
\newblock In \emph{Proceedings of the 2018 Conference on Empirical Methods in Natural Language Processing}, pages 2369--2380, Brussels, Belgium. Association for Computational Linguistics.

\bibitem[{Zeng et~al.(2025)Zeng, Huang, Liu, Liu, He, Ma, and He}]{zeng2025simplerlzooinvestigatingtamingzero}
Weihao Zeng, Yuzhen Huang, Qian Liu, Wei Liu, Keqing He, Zejun Ma, and Junxian He. 2025.
\newblock \href {https://arxiv.org/abs/2503.18892} {Simplerl-zoo: Investigating and taming zero reinforcement learning for open base models in the wild}.
\newblock \emph{Preprint}, arXiv:2503.18892.

\bibitem[{Zhang and Zuo(2025)}]{zhang2025grpoleaddifficultyawarereinforcementlearning}
Jixiao Zhang and Chunsheng Zuo. 2025.
\newblock \href {https://arxiv.org/abs/2504.09696} {Grpo-lead: A difficulty-aware reinforcement learning approach for concise mathematical reasoning in language models}.
\newblock \emph{Preprint}, arXiv:2504.09696.

\bibitem[{Zhang et~al.(2025{\natexlab{a}})Zhang, Yang, Gao, Chen, Hu, Chen, Wang, Guo, Zheng, Wang, and Zhao}]{zhang2025letslearningthinkandsearchprocessandoutcome}
Qi~Zhang, Shouqing Yang, Lirong Gao, Hao Chen, Xiaomeng Hu, Jinglei Chen, Jiexiang Wang, Sheng Guo, Bo~Zheng, Haobo Wang, and Junbo Zhao. 2025{\natexlab{a}}.
\newblock \href {https://arxiv.org/abs/2505.17447} {Lets: Learning to think-and-search via process-and-outcome reward hybridization}.
\newblock \emph{Preprint}, arXiv:2505.17447.

\bibitem[{Zhang et~al.(2025{\natexlab{b}})Zhang, Li, Dong, Wang, Jia, Li, Zhang, Xu, Du, Guo, Tang, and Zhao}]{zhang2025processvsoutcomereward}
Wenlin Zhang, Xiangyang Li, Kuicai Dong, Yichao Wang, Pengyue Jia, Xiaopeng Li, Yingyi Zhang, Derong Xu, Zhaocheng Du, Huifeng Guo, Ruiming Tang, and Xiangyu Zhao. 2025{\natexlab{b}}.
\newblock \href {https://arxiv.org/abs/2505.14069} {Process vs. outcome reward: Which is better for agentic rag reinforcement learning}.
\newblock \emph{Preprint}, arXiv:2505.14069.

\bibitem[{Zhang et~al.(2024)Zhang, Wang, Yang, Wang, Feng, and Zhang}]{zhang2024hierarchicalretrievalaugmentedgenerationmodel}
Xiaoming Zhang, Ming Wang, Xiaocui Yang, Daling Wang, Shi Feng, and Yifei Zhang. 2024.
\newblock \href {https://arxiv.org/abs/2408.11875} {Hierarchical retrieval-augmented generation model with rethink for multi-hop question answering}.
\newblock \emph{Preprint}, arXiv:2408.11875.

\bibitem[{Zhao et~al.(2025)Zhao, Wang, Xu, Zha, and Liu}]{zhao2025rsearchempoweringllmreasoning}
Qingfei Zhao, Ruobing Wang, Dingling Xu, Daren Zha, and Limin Liu. 2025.
\newblock \href {https://arxiv.org/abs/2506.04185} {R-search: Empowering llm reasoning with search via multi-reward reinforcement learning}.
\newblock \emph{Preprint}, arXiv:2506.04185.

\bibitem[{Zhou et~al.(2024)Zhou, Liu, Jin, Nie, and Dou}]{zhou2024metacognitive}
Yujia Zhou, Zheng Liu, Jiajie Jin, Jian-Yun Nie, and Zhicheng Dou. 2024.
\newblock \href {https://openreview.net/forum?id=TW2gJyR6Mj} {Metacognitive retrieval-augmented large language models}.
\newblock In \emph{The Web Conference 2024}.

\end{thebibliography}
\clearpage
\appendix
\section{Related Work}
\label{sec:rel}
\noindent \textbf{Retrieval-Augmented Generation. }
\vic{
RAG has emerged as a prominent framework to augment LLMs with external knowledge, aiming to mitigate issues such as hallucination \cite{ayala-bechard-2024-reducing}, domain incompleteness \cite{siriwardhana-etal-2023-improving}, and temporal staleness \cite{gade2024itstimeincorporatingtemporality}. Conventional RAG systems adopt a static retrieve-then-generate paradigm, where a retriever first fetches top-ranked documents given an input query, and a generator conditions its output on the retrieved context \cite{NEURIPS2020_6b493230,10.5555/3524938.3525306}. While this structure has proven effective for factual QA and open-domain generation, it often falls short when confronted with complex reasoning tasks involving multi-hop dependencies \cite{tang2024multihoprag}, latent constraints \cite{li2024elicitreasoningllmscriticguided}, or ambiguous query intents \cite{chan2024rqrag}. To overcome these limitations, recent works have proposed more cognitively informed RAG architectures. For example, AdaptiveRAG~\cite{jeong-etal-2024-adaptive} uses query classification to trigger different retrieval strategies, PlanRAG~\cite{lee-etal-2024-planrag} decomposes tasks into executable plans for targeted retrieval, and ITER-RETGEN~\cite{shao-etal-2023-enhancing} incorporates intermediate generation to iteratively reformulate queries. In parallel, modular and hybrid RAG frameworks \cite{gao2024modularragtransformingrag,zhang2024hierarchicalretrievalaugmentedgenerationmodel,zhou2024metacognitive}  have introduced componentized systems that integrate query rewriting, evidence aggregation, and verification in sequential or recursive pipelines. These advances suggest that effective RAG increasingly requires not just better retrieval quality, but dynamic, context-aware reasoning to inform when, what, and how to retrieve. This growing interdependence between retrieval and reasoning sets the stage for more adaptive mechanisms (particularly those that go beyond pre-defined rules) highlighting the need for learning-based control in complex RAG workflows.}

\smallskip \noindent \textbf{Reinforcement Learning for LLM Reasoning. }
\vic{Driven by the growing need for LLMs to perform complex and reliable reasoning across diverse tasks, recent research has turned to reinforcement learning  as a promising paradigm to enhance their reasoning capabilities. The release of GPT-o1~\cite{openai2024openaio1card} and DeepSeek-R1~\cite{deepseekai2025deepseekr1incentivizingreasoningcapability} marked a shift toward training LLMs that exhibit structured, multi-step reasoning through RL-based objectives. Early efforts such as SimpleRL-Zoo~\cite{zeng2025simplerlzooinvestigatingtamingzero} and Open-Reasoner-Zero~\cite{hu2025openreasonerzeroopensourceapproach} explored direct RL fine-tuning from base models, eliminating the reliance on extensive supervised instruction tuning. Building on this foundation, approaches like DeepScaler~\cite{10.1109/ASE56229.2023.00038} and Light-R1~\cite{wen2025lightr1curriculumsftdpo} introduced cold-start datasets and reward schemes explicitly designed to promote step-by-step thinking and verifiable inferences. In parallel, improvements to RL algorithms, such as Dr GRPO~\cite{liu2025understandingr1zeroliketrainingcritical}, refined policy optimization to better align with the cognitive demands of long-form reasoning.}

\smallskip \noindent \textbf{Reinforcement Learning for RAG Reasoning. }
\vic{There is a growing body of work focused on bringing RL-based reasoning into the retrieval-augmented generation framework \cite{song2025r1searcherincentivizingsearchcapability,jin2025searchr1trainingllmsreason,chen2025researchlearningreasonsearch}. Inspired by DeepSeek-R1, these approaches use regularized rewards to encourage the model to think and generate retrieval queries to search external corpora. However, these methods only consider the final outcome reward signal to train the model, which is relatively simple, and do not deeply optimize according to the characteristics of the RAG task itself. In view of this, many works incorporate the model’s thinking process into the reward calculation \cite{wang2025stepsearchignitingllmssearch,sha2025semreinforcementlearningsearchefficient,zhang2025processvsoutcomereward}. For example, \citet{zhao2025rsearchempoweringllmreasoning} uses different models to generate answers based on the same evidence in order to calculate evidence quality, while also alleviating answer bias caused by the preference of the policy model itself. R3-RAG \cite{li2025r3raglearningstepbystepreasoning} calculates the relevance between each retrieved document and the question at every step, attempting to improve the model’s search strategy through fine-grained process rewards. 
Different from these works, our work directly measures the sufficiency between all retrieved documents and the question, in order to enhance the model’s awareness of searching less or more. In addition, some works attempt to add new information fields to the “search–thinking–answer” pipeline proposed by search-r1-type methods to prompt the model to think more about the documents \cite{ren2025effectivetransparentragadaptivereward}. For example, \citet{shi2025searchrefinethinkautonomous} lets the model refine the retrieved documents during the reasoning process before proceeding to thinking. Our work, unlike in these methods, directly rewards the model’s thinking process, allowing the model to optimize its reasoning process, rather than manually adding thinking rules.}

\section{Experimental Setup for Preliminary Studies}
\vic{We trained a reasoning‑capable RAG model using the GRPO algorithm from Deepseek-R1. Following R1‑Searcher, we selected part of the training data from HotpotQA \cite{yang-etal-2018-hotpotqa} and 2WikiMultiHopQA \cite{ho-etal-2020-constructing}, with a total of 8,148 examples. We used the Qwen‑2.5‑3B‑Instruct model for training. For the retriever, we deployed a system locally based on the BGE-large-en-v1.5 retrieval, with the retrieval corpus from English Wikipedia provided by KILT \cite{petroni-etal-2021-kilt} in 2019. The training prompt can be found in Appendix~\ref{prompt_tem}. We trained the model for one epoch (detailed experimental settings are the same as those in Section~\ref{sec:exp}.) and evaluated the model on the in‑domain 2Wiki test set and the out‑of‑domain Musique test set. After obtaining predictions on 500 test examples from each dataset, we used GPT‑4o to evaluate the correlation between the model’s predictions and the retrieved content plus the reasoning process. } 

\vic{We introduce notation for a generic open-domain question–reasoning setting, assuming sufficient contextual information.
Consider an example represented as $E = (Q, RD, A)$, where $Q$ is the query, $RD$ is the combined reasoning process and retrieved documents, and $A$ is either the model’s predicted answer or the gold answer. We define $RD = \{R_1, D_1, \dots, R_i, D_i, R_{i+1}\}_{i=1}^n$, where $n$ is the number of sub-questions generated by the model during reasoning, $R_i$ denotes the intermediate reasoning step prior to document retrieval, and $D_i$ represents the set of documents retrieved in response to the sub-question formulated in $R_i$.}

\vic{ We define three cases:  (1) \textbf{Overthinking}: the model produces too many reasoning steps, meaning that the gold answer could already be inferred at some step $R_i$ with $i < n$.  
(2) \textbf{Good thinking}: the model obtains sufficient content exactly before the final reasoning step, i.e., the gold answer can only be inferred at step $R_{n+1}$.  
(3) \textbf{Underthinking}: the model stops reasoning without obtaining sufficient content in $RD$, and still outputs an answer.}
\vic{Since \citet{joren2025sufficient} has shown    enabling LLMs to judge whether the provided context is adequate for answering a question is effective, we directly input $Q$, $RD$, and the gold answer into an LLM to assess whether $Q$ and $RD$ together are sufficient to derive the correct answer (see prompt in App.\ref{prompt_tem}).
Importantly, we include the gold answer in the input, unlike what \cite{joren2025sufficient} does. 
 which uses only $Q$ and $RD$. This is because we found that omitting the answer leads to more errors: when $RD$ is actually insufficient, the model is more likely to incorrectly judge it as sufficient.}

\section{Implementation Details and Baselines}
\label{app_ex_details}

\subsection{Implementation Details.}
\vic{For all baselines and \textbf{TIRESRAG-R1}, we use Qwen-2.5-3B (base and instruct variants) as the backbone.  Following R1-Searcher, we use the 2019 English Wikipedia as the external knowledge source, and BGE-large-en-v1.5 as the retrieval engine.  
Our training framework is built upon OpenRLHF \cite{hu2025openrlhfeasytousescalablehighperformance}, and we use FlashRAG \cite{10.1145/3701716.3715313} for evaluation.  
The total batch size is 108, with a learning rate of $2\times10^{-6}$.  
We generate 5 rollout samples per input for reward estimation and set temperature = 1 during rollout to encourage exploration.}  
\vic{The KL coefficient $\beta$ is set to 0, the number of iterations per batch $\mu$ = 2, and the clipping parameter $\epsilon = 0.2$.  
Top-$k$ during retrieval is set to 5.   In the difficulty-aware resampling strategy, the hyperparameters are set to $A = 0.4$, $B = 1.5$, $\rho_0 = 0.75$, and $k = 10.0$.  
For the consistency penalty, we set $\lambda_p = 0.1$.  
In the overall reward computation, the weights are configured as $w_t = 0.6$, $w_s = 0.3$, and $w_r = 0.3$.
All models are trained on 3 NVIDIA H200 GPUs: 2 GPUs are allocated for policy optimization, and 1 GPU is dedicated to rollout inference via vLLM~\cite{10.1145/3600006.3613165}.}

\subsection{Evaluation Metrics}
\label{eva_metrcis}
\vic{For open-ended evaluation, we report three widely used metrics: \textbf{Exact Match (EM)}, \textbf{F1 score} (aligned with the RL training answer reward), and \textbf{LLM-as-Judge}.  
EM measures whether the ground-truth answer exactly matches the model prediction.  
F1 is computed between predicted and gold answers to handle partial overlap.  
For LLM-as-Judge, we use GPT-4o to evaluate the semantic correctness of the prediction given the question and supporting evidence.  
The prompts used for LLM-as-Judge are listed in App.~\ref{prompt_tem}.
Due to the high cost of using LLM-as-a-Judge, we only adopt this metric in the main experiments. For the other experiments, we use \textbf{Cover Exact Match (CEM)}, which assesses whether the ground-truth answer is included in the predicted answer.}

\subsection{Baselines.} 
\label{sec:app_baseline}
To evaluate the effectiveness of our proposed \textbf{TIRESRAG-R1} method, we implement and compare against 14 baselines, grouped into four categories:

\textit{(1) Naive prompt methods:}  
\textbf{Direct} answers questions directly using its own parametric knowledge.  
\textbf{COT} is instructed to produce a chain of thought before the final answer.  
\textbf{R1-based} uses a distilled DeepSeek-Qwen-3B reasoning model to first reason then answer.  
We implement all three naive prompt methods ourselves.

\textit{(2) Retrieval-augmented prompt methods:}  
\textbf{Naive RAG} extends \textbf{Direct} by retrieving documents for the query and appending them as additional input before direct answering.  
\textbf{Agentic-R1} \cite{li2025searcho1agenticsearchenhancedlarge} enables the model to autonomously retrieve external knowledge when needed while avoiding interference from irrelevant content.  
\textbf{Search-o1} \cite{li2025searcho1agenticsearchenhancedlarge} introduces a Reason-in-Documents module that condenses retrieved content into coherent reasoning steps, iteratively guiding the model to the final answer.  
\textbf{SURE} \cite{kim2024sure} generates and evaluates summaries of retrieved passages for multiple answer candidates.  
\textbf{IRCOT} \cite{trivedi-etal-2023-interleaving} interleaves retrieval with Chain-of-Thought reasoning.  
\textbf{Self-Ask} \cite{press-etal-2023-measuring} improves multi-hop reasoning by decomposing the original question into intermediate sub-questions that are answered before final prediction.  
\textbf{RQRAG} \cite{chan2024rqrag} learns to explicitly refine queries through rewriting, decomposition, and disambiguation.  
For these retrieval-augmented methods, we use the implementations provided in \texttt{flashrag}.

\textit{(3) SFT methods:}  
\textbf{SFT} fine-tunes the model directly on training pairs of questions and gold answers. During training, we only input the question and instruct the model to output a short answer without any intermediate reasoning or evidence selection.  
\textbf{SimpleDeepSearcher} \cite{sun2025simpledeepsearcherdeepinformationseeking} constructs a high-quality dataset containing intermediate reasoning and retrieval steps, then fine-tunes the model with question-to-(reasoning,answer) pairs. Its input is also only the question, but its output format follows R1-Searcher with four components:
\begin{itemize}
\item <thinking> ... </thinking>,
\item <|begin\_search\_query|> ... </|end\_search\_query|>,
\item <|begin\_search\_result|> ... </|end\_search\_result|>,
\item \textbackslash box\{answer\}.
\end{itemize}
A total of 871 examples are constructed to reproduce this method, and during inference we extract the text inside the \textbackslash box\{\} as the model’s predicted answer.

\textit{(4) RL methods:}  
\textbf{Search-R1} \cite{jin2025searchr1trainingllmsreason} uses only F1 as reward.  
\textbf{R1-Searcher} \cite{song2025r1searcherincentivizingsearchcapability} uses answer reward plus a format reward.  
\textbf{Research} \cite{chen2025researchlearningreasonsearch} also incorporates format reward.  
To ensure fairness, we re-trained these RL-based baselines on our training set using the authors’ released code and hyperparameter settings, without using their checkpoints.  
\textbf{LeTS} \cite{zhang2025letslearningthinkandsearchprocessandoutcome} combines step-level rewards with answer rewards and introduces rollout-level redundancy penalties and group-level knowledge-matching rewards. Since LeTS has no public code, we report the results from the original paper.

\smallskip
All baselines are evaluated under a unified protocol: each method first produces its predicted answers, which are then scored using the same standardized evaluation script.

\begin{figure*}[!htbp]
    \centering
    \includegraphics[width=1\linewidth]{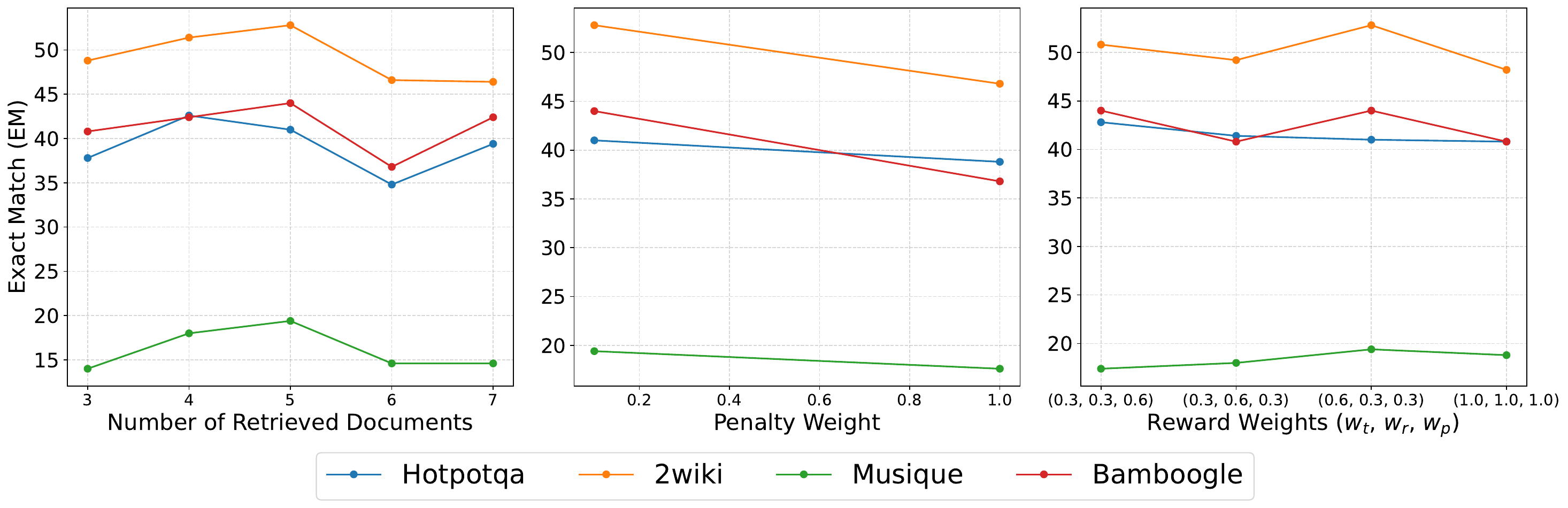}
    \caption{Effect of different hyperparameters.}
    \label{fig:hyper}
\end{figure*}

\section{Detailed Reinforce++}
\label{app_reinforce}
\vic{Reinforce++~\cite{hu2025reinforceefficientrlhfalgorithm} is an efficient RLHF algorithm without a critic network, designed to address overfitting in advantage estimation and reward hacking in REINFORCE-based methods.  
Its core idea is to use the global batch mean reward as the baseline, rather than constructing a separate baseline for each prompt as in RLOO or GRPO. This avoids prompt-specific bias and improves stability and generalization.  
We adopt the same reward computation strategy as in our main experiments to ensure consistency across training and evaluation.}

\section{Visualization of Annealing}
\begin{figure}
    \centering
    \includegraphics[width=1\linewidth]{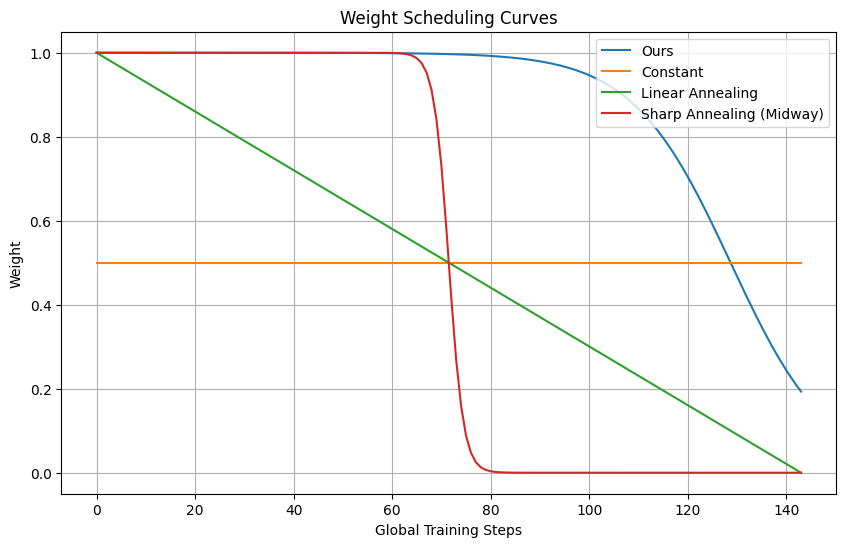}
    \caption{Visualization of different reward weight
scheduling strategies curves.}
    \label{fig:linear}
\end{figure}
\label{vis_anneal}
\vic{To better illustrate the dynamic weight schedule defined in the main content, Figure~\ref{fig:linear} plots the four annealing strategies used in Section~7.2.
Our strategy keeps the weight stable in the early and mid training phases, and then drops sharply in the late phase, aligning with our goal of letting the model focus on answer accuracy in the later stages.}

\begin{table*}
\footnotesize
    \begin{tabular}{l|cccccccccccc}
\toprule
Method & \multicolumn{3}{c}{Hotpotqa} & \multicolumn{3}{c}{2wikimultihopqa} & \multicolumn{3}{c}{Musique} & \multicolumn{3}{c}{Bamboogle} \\
 & EM & F1 & CEM & EM & F1 & CEM & EM & F1 & ACC & CEM & F1 & CEM \\
\midrule
Naive GRPO & 36.4 & 48.5 & 41.6 & 46.4 & 53.1 & 54.0 & 16.2 & 26.0 & 19.6 & 36.0 & 49.2 & 40.8 \\
\rowcolor{yellow!20}

TIRESRAG-R1-Instruct & \textbf{41.0} & \textbf{54.2} & \textbf{46.0} & \textbf{52.8} & \textbf{59.6} & \textbf{60.8} & \textbf{19.4} & \textbf{30.0} & \textbf{23.2} & \textbf{44.0} & \textbf{54.7} & \textbf{47.2} \\
 w/o Filter & 18.8 & 24.6 & 26.8 & 21.4 & 26.5 & 28.8 & 6.0 & 11.4 & 9.4 & 19.2 & 28.8 & 25.6 \\

w/o Difficulty & 38.2 & 50.4 & 43.6 & 49.2 & 54.0 & 55.4 & 17.0 & 27.0 & 21.4 & 35.2 & 49.3 & 38.4 \\
w/o Penalty & 38.0 & 49.9 & 44.2 & 44.2 & 50.9 & 52.0 & 15.6 & 24.9 & 19.0 & 39.2 & 50.6 & 43.2 \\
 w/o Reflect & 37.8 & 47.9 & 41.2 & 44.6 & 52.0 & 53.8 & 10.8 & 21.4 & 15.2 & 32.8 & 45.3 & 37.6 \\
w/o Sufficient & 37.4 & 48.5 & 44.6 & 41.4 & 46.3 & 46.4 & 14.0 & 22.3 & 16.4 & 32.6 & 43.4 & 35.8 \\
w/o Thinking & 39.8 & 51.8 & 47.6 & 44.8 & 51.3 & 53.8 & 14.8 & 23.9 & 19.2 & 37.6 & 46.4 & 39.2 \\

\bottomrule
\end{tabular}
\caption{Ablation Study.}
\label{tab:abla}
\end{table*}

\section{Ablation Study}
\vic{To verify the effectiveness of each module in \textbf{TIRESRAG-R1}, we conduct systematic ablation experiments on the Qwen2.5-3B-Instruct model. Table~\ref{tab:abla} shows the obtained results. We observe that:  
\textbf{(1)} The \textbf{filter module} is crucial for model stability. As shown in Table~\ref{tab:abla}, removing it significantly degrades performance: compared with standard GRPO, the average F1 score drops by 21.2\%. This is because, in the later training stages, the model collapses. We show the training curves in Section~7.3.  
\textbf{(2)} The \textbf{difficulty-aware weighting} and \textbf{penalty} mechanisms are crucial for effectively integrating our diverse reward signals. While ablations without them still outperform naive GRPO, the gains are limited, showing average performance drops of 4.48 and 5.58 points compared to our full method.  }
\textbf{(3)} \vic{Each reward component plays a critical role in overall performance. Removing any single reward leads to noticeable degradation—and in all cases, performance drops below that of naive GRPO. Notably, removing the \textbf{sufficient reward} results in the largest decline, indicating that the model may engage in reward hacking and neglect crucial external documents. In contrast, removing the \textbf{thinking reward} has the smallest impact, with an average drop of 6.3\% compared to our full method. This is likely because the \textbf{answer}, \textbf{sufficient}, and \textbf{reflect} rewards already provide partial supervision for generating high-quality reasoning chain.}

\section{Analysis}
\label{sec:analy}
\subsection{Impact of Hyperparameters. }
\vic{We explore three key hyperparameters: the number of retrieved documents, the penalty weight, and the reward mixture weights. Figure~\ref{fig:hyper} present the obtained  results.}
\begin{itemize}[itemsep=0.5ex, leftmargin=3mm]
\item 
\textit{Impact of number of retrieved documents.}  
\vic{As the number of retrieved documents increases from 3 to 5 (ours), the model performance improves, with EM increasing by 3.95 points on average. When increasing from 5 (ours) to 7, the average EM drops by 3.6 points, suggesting that retrieving too many documents introduces noise, hurting reasoning quality. There is thus a trade-off between providing more information and avoiding noise.}

\item \textit{Impact of penalty weight.}  
\vic{In Eq.~\ref{con-pen}, $\lambda_p$ controls the magnitude of the consistency penalty in the advantage calculation. Setting $\lambda_p=0.5$ or $\lambda_p=1$, we observe that as $\lambda_p$ increases, performance decreases, with the largest drop when $\lambda_p=1$. This suggests over-emphasizing consistency can suppress beneficial reward signals and hurt final performance.}

\item \textit{Impact of reward mixture weights.}  
\vic{In Eq.~\ref{reward_sum}, thinking, sufficient, and reflect rewards are combined with different weights. We try settings of (0.3,0.3,0.6), (0.3,0.6,0.3), (0.3,0.3,0.3), and our (0.6,0.3,0.3). Results show that giving the highest weight to thinking reward (ours) yields the best performance, likely because thinking directly measures the quality of the reasoning chain, which more strongly impacts final answers.}
\end{itemize}
\begin{figure}
    \centering
    \includegraphics[width=0.95\linewidth]{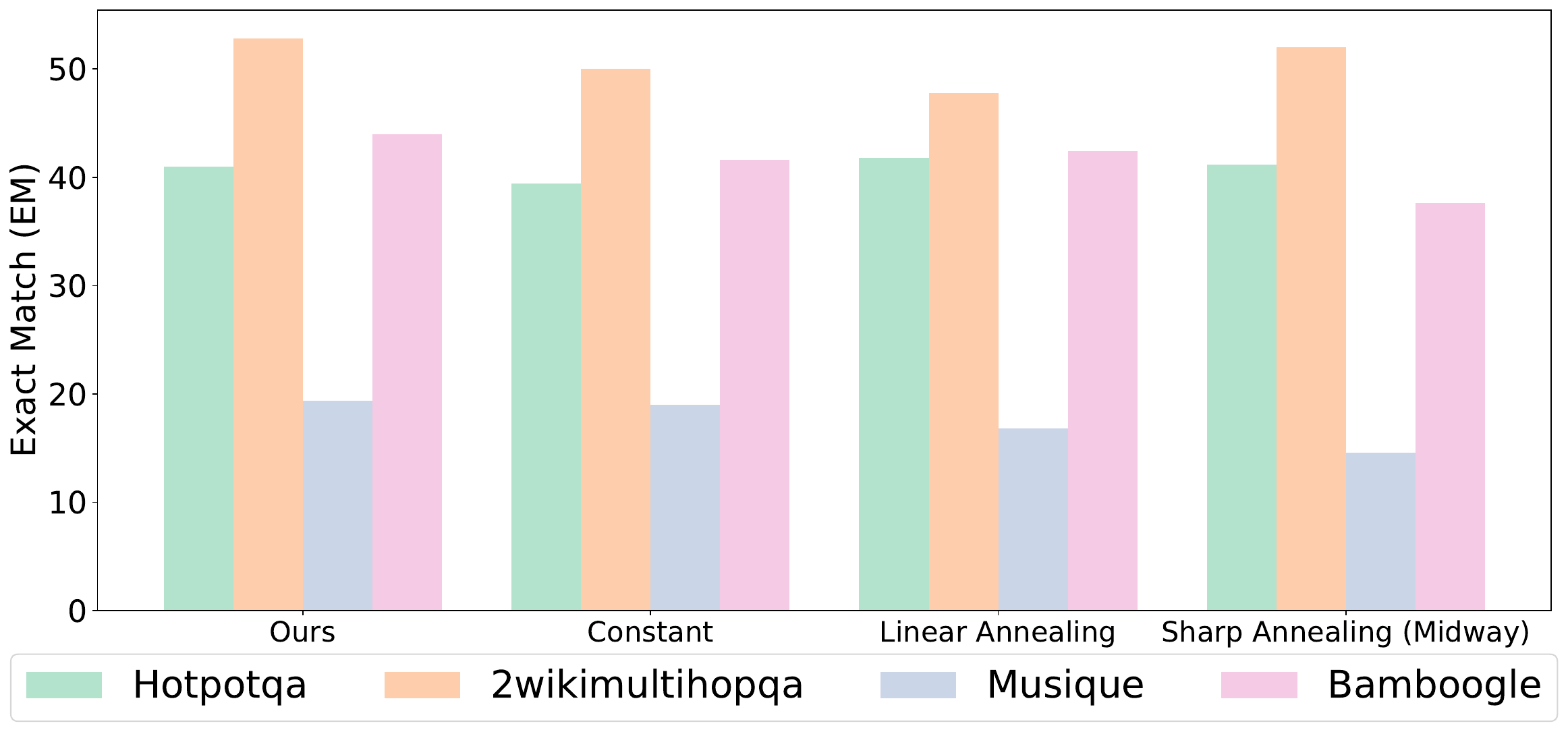}
    \caption{Comparison of different reward weight scheduling strategies.}
    \label{fig:res_annealing}
\end{figure}

\subsection{Impact of Annealing Strategy. }
\vic{As described in Section~\ref{sec:reward}, we adopt a decaying schedule for mixed reward weights over training steps. To analyze its effectiveness, we compare several variants (fixed weights, linear decay, fast decay). Weight curves are shown in Fig.~\ref{fig:linear}. Results in Fig.~\ref{fig:res_annealing} show that linear annealing performs worst, with an average EM drop of 2.35 points 
across datasets, suggesting auxiliary signals decay too early.  
In contrast, our proposed schedule achieves the best or second-best results on all datasets, especially on 2WikiMultiHopQA and Bamboogle. Fixed scheduling is relatively stable but suboptimal. Pre-low annealing performs slightly better on HotpotQA but worse on others.}
\begin{figure}
    \centering
    \includegraphics[width=1\linewidth]{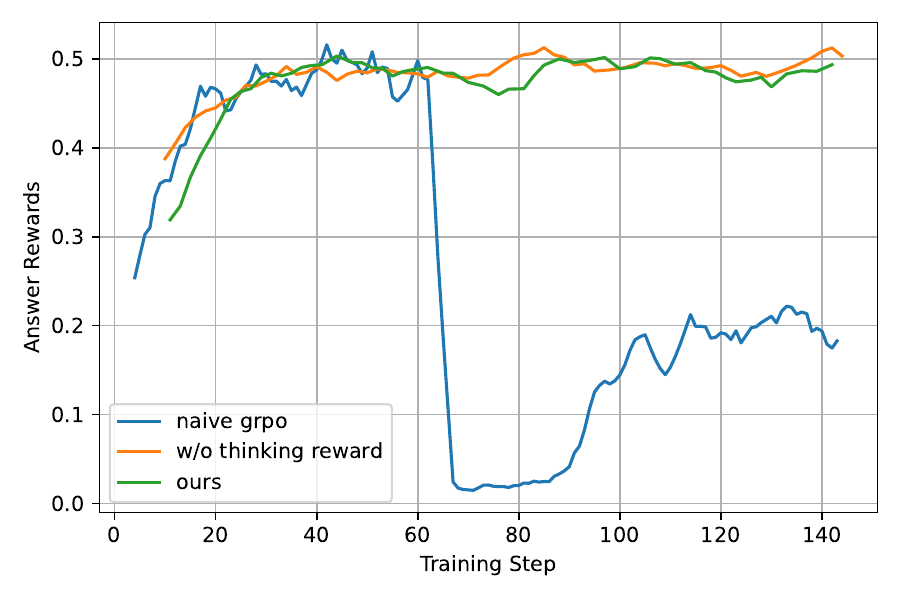}
    \caption{Training dynamics of answer rewards over steps.}
    \label{fig:answer_reward}
\end{figure}
\begin{figure}
    \centering
    \includegraphics[width=1\linewidth]{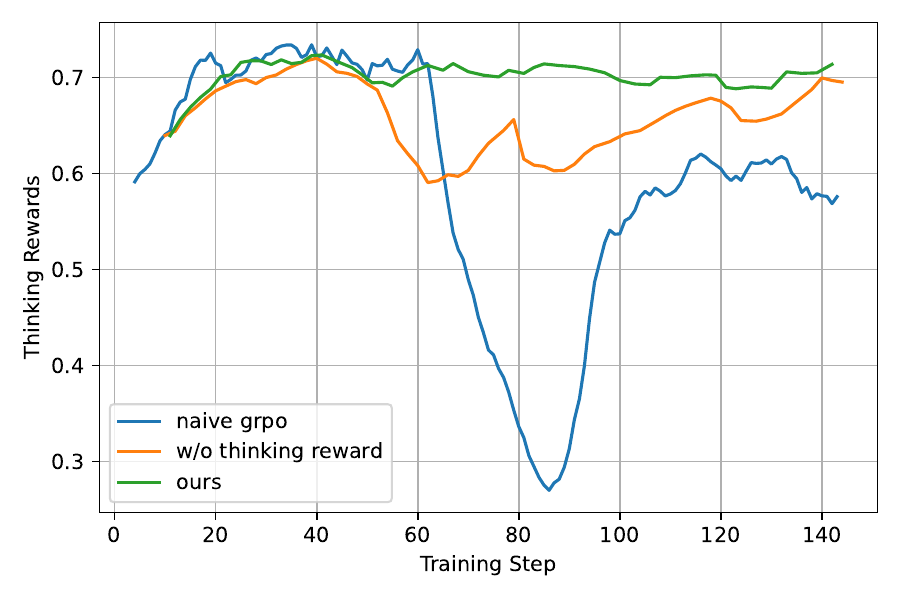}
    \caption{Training dynamics of thinking rewards over steps.}
    \label{fig:thinking_reward}
\end{figure}

\begin{table*}
\centering
\footnotesize
\begin{tabular}{l|ccccccccc}
\toprule
Method & \multicolumn{3}{c}{Nq} & \multicolumn{3}{c}{Popqa} & \multicolumn{3}{c}{Triviaqa} \\
 & EM & F1 & CEM & EM & F1 & CEM & EM & F1 & CEM \\
\midrule
Direct Generation & 6.8 & 10.8 & 9.5 & 8.6 & 11.8 & 9.4 & 7.5 & 19.0 & 9.5 \\
COT & 10.5 & 17.5 & 15.0 & 9.7 & 13.8 & 10.7 & 7.8 & 20.6 & 10.1 \\
\hline 
Naive RAG & 23.4 & 32.9 & 31.9 & 29.4 & 37.0 & 34.5 & 12.9 & 29.9 & 16.5 \\
Sure & 25.5 & 34.2 & 27.9 & 30.4 & 35.7 & 31.1 & 13.4 & 29.7 & 16.0 \\
IRCOT & 15.4 & 25.1 & 33.2 & 25.1 & 31.5 & 35.8 & 10.1 & 25.0 & 17.6 \\
Self-ask & 17.2 & 27.1 & 41.4 & 24.8 & 31.7 & 41.5 & 9.5 & 24.1 & 20.4 \\
\hline 

SFT & 6.3 & 12.9 & 10.1 & 7.6 & 11.8 & 8.4 & 5.2 & 14.9 & 7.0 \\
SimpleDeepSearcher & 33.4 & 44.0 & 44.1 & 38.9 & 44.3 & 44.1 & 59.6 & 67.2 & 66.5 \\
\hline 
ReSearch-Instruct & 35.8 & 46.2 & 44.5 & 41.8 & 47.4 & 46.3 & 58.4 & 66.1 & 64.1 \\
Search-R1-Instruct & 34.2 & 44.0 & 44.2 & 37.9 & 43.5 & 44.1 & 54.5 & 62.2 & 62.3 \\
R1-search-Instruct & 35.2 & 46.4 & 44.8 & 40.3 & 46.1 & 45.6 & 57.3 & 65.6 & 64.0 \\
\hline 
\rowcolor{yellow!20}
\textbf{TIRESRAG-R1-Instruct} & \textbf{38.0} & \textbf{49.1} & \textbf{47.9} & \textbf{43.0} & \textbf{48.8} & \textbf{47.9} & \textbf{60.0} & \textbf{68.2} & \textbf{66.9} \\
\bottomrule
\end{tabular}
\caption{Generalization results on single-hop benchmarks (NQ, PopQA, TriviaQA).}
\label{tab:single_hop}
\end{table*}
\begin{figure}
    \centering
    \includegraphics[width=0.95\linewidth]{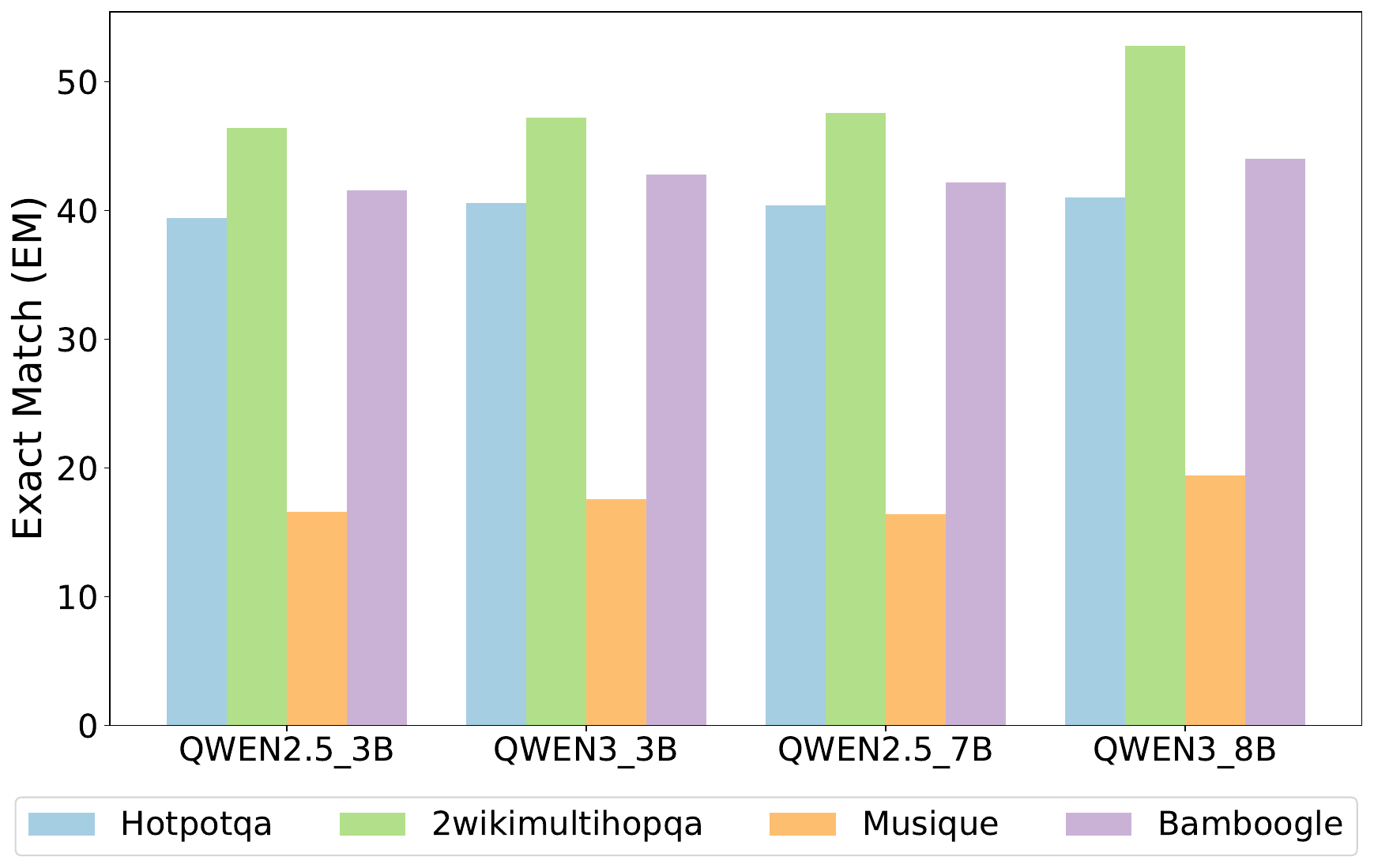}
    \caption{Comparison of different reward backbone models.}
    \label{fig:res_reward_backbone}
\end{figure}

\subsection{Learning Curve Analysis. }
\vic{Figures~\ref{fig:answer_reward} and \ref{fig:thinking_reward} show training dynamics. For naive GRPO, because it does not filter out ``all-correct'' or ``all-wrong'' queries, its answer reward is similar to ours early on but then drops sharply, showing clear collapse, before slowly recovering but remaining below initial levels.  
Our method maintains stability throughout training, with answer reward staying high.  
For the baseline without thinking reward, answer reward remains similar to ours, but thinking reward fluctuates and declines, indicating the model fails to learn a stable reasoning process. This further shows that process-level rewards help both reasoning quality and training stability.}
\subsection{Influence of Reward Backbone Model. }
\vic{The reward model plays a critical role. Fig.~\ref{fig:res_reward_backbone} compares using Qwen2.5-3B and Qwen3-3B for sufficient and thinking scoring.  
We observe consistent improvements with Qwen3-3B: on Musique, EM improves from 16.6 to 17.6; on Bamboogle, EM improves from 41.6 to 42.8.  
A similar trend is seen comparing Qwen2.5-3B and Qwen2.5-7B: on HotpotQA, EM improves from 39.4 to 40.4 and F1 from 52.7 to 54.4.  
These results show that stronger reward models provide better supervision and thus improve final performance.}

\section{Generalization on Single-Hop 
Benchmarks}
\label{sec:app_gen_single}
We present in Table~\ref{tab:single_hop} the complete experimental results with more baselines and additional metrics. As shown, our method outperforms all baselines across all metrics.

\section{Case Study}
\vic{Figures~\ref{fig:case1}, \ref{fig:case2}, and \ref{fig:case3} present three examples illustrating the effectiveness of TIRESRAG-R1.  
In the first example, the TIRESRAG-R1-trained Qwen2.5-3B follows a correct reasoning process and gives the correct answer, while the naive GRPO model, despite having sufficient information, produces an incorrect reasoning chain and thus a wrong answer.  
In the second example, the TIRESRAG-R1-trained model successfully retrieves enough information before answering, whereas the naive GRPO model guesses prematurely without sufficient evidence and produces a wrong answer.  
In the third example, the naive GRPO model has sufficient information and correct reasoning but still outputs the wrong answer, while our model successfully produces the correct answer.  
These examples show that TIRESRAG-R1 guides models toward better reasoning chains and ensures stronger consistency between reasoning and answers.}

\section{Prompt Templates}
\label{prompt_tem}
\vic{
Figures~\ref{prompt_rag_reasoning_instruct} to~\ref{prompt_llm_as_judge} present all prompt templates mentioned in this paper.
\begin{itemize}
\item
\textbf{Figure~\ref{prompt_rag_reasoning_instruct}:} Prompt used for training instruction-based models, setting the system’s reasoning–retrieval–reflection strategy and output format.
\item
\textbf{Figure~\ref{prompt_rag_reasoning_base}:} Prompt used for training base models, converting the above instruction-based prompt into a conversation format.
\item
\textbf{Figure~\ref{prompt_suff_reward}:} Prompt for scoring the sufficiency of reasoning trajectories. We follow the sufficient-context method: the model is prompted to list sub-questions that solve the main question, and only if all sub-questions can be answered from the given references is the trajectory judged sufficient. A demonstration example is provided.
\item
\textbf{Figure~\ref{prompt_think_reward}:} Prompt for scoring reasoning quality. The four criteria described in the main text (logical soundness, alignment, error awareness, conciseness) are included. The model is prompted to output only a number between 0 and 1 to avoid format errors. In experiments, the reward model followed the prompt strictly and output valid scores.
\item
\textbf{Figure~\ref{prompt_llm_as_judge}:} Prompt used for LLM-as-Judge evaluation. We embed both the predicted answer and the gold answer into the prompt and feed it to GPT-4o. If GPT-4o judges that the predicted and gold answers are semantically equivalent, it returns “yes,” otherwise “no.”
\end{itemize}}
\section{Algorithm}
\label{app_algorithm}
\vic{The pseudo code for TIRESRAG-R1  is shown in Algorithm \ref{alg:lenrag}.}
\begin{algorithm*}[t]
\small
\caption{TIRESRAG-R1 Training with GRPO and Multi-Dimensional Rewards}
\label{alg:lenrag}
\begin{algorithmic}[1]
\REQUIRE Policy model $\pi_\theta$, reference model $\pi_{\theta_{\text{old}}}$, dataset $\mathcal{D}$, retrieval model $\pi_{\text{ret}}$, hyperparameter weights $w_{\text{think}}, w_{\text{suff}}, w_{\text{reflect}}$, KL penalty $\beta$, iterations $T$, rollouts per query $G$, buffer $B$, dynamic weight $d_w$
\ENSURE Updated policy model $\pi_\theta$
\FOR{$t = 1, \dots, T$}
  \STATE Compute dynamic weight:
  \[
    a_t \gets \frac{1}{1 + \exp\!\left(\frac{T - 0.9t}{10}\right)}
  \]
  \STATE Sample $Q \subset \mathcal{D}$
  \FOR{$q \in Q$}
    \STATE Generate rollouts $\{y_i\}_{i=1}^G \sim \pi_\theta(\cdot \mid q)$
    \FOR{$i = 1, \dots, G$}
      \STATE Extract reasoning trajectory $RD_i$ and prediction $a_i$ from $y_i$
      \STATE Compute answer reward:
      \[
        R_i^{A} \gets \mathrm{F1}(a_i, a^*)
      \]
      where $a^*$ is the gold answer for $q$
      \STATE Compute sufficient reward:
      \[
      \mathrm{R^S_i}(q,RD_i,a^*)=
      \begin{cases}
      1, & RD_i \text{ contains sufficient info to derive } a^*,\\
      0, & \text{otherwise}
      \end{cases}i
      \]
      \STATE Compute think reward:
      \[
        R_i^{T} \gets \mathrm{Think}(RD_i), \quad \mathrm{Think}(RD_i)\in[0,1]
      \]
      scored on logic, alignment, error-awareness, and conciseness
      \STATE Compute reflection reward:
      \[
      \mathrm{R^T_i}(a_i)=
      \begin{cases}
      +1,& \text{if } CEM(a_1, a^*) = 0  \text{ and }  CEM(a_2, a^*) = 1, \\
      -1,& \text{if } CEM(a_1, a^*) = 1 \text{ and } CEM(a_2, a^*) = 0,\\ 

      0,& \text{otherwise}
      \end{cases}
      \]
      \STATE Combine rewards:
      \[
      R^{sum}_i = a_t \cdot \bigl(w_{\text{think}} R_i^{T} + w_{\text{suff}} R_i^{S} + w_{\text{reflect}} R_i^{R}\bigr) + R_i^{A}
      \]
    \ENDFOR

    \IF{$0.1 < \forall\ r_i^{(a)} < 0.9$ for  $i \in \{1,\dots,G\}$}
      \STATE Compute advantage by normalizing batch reward:
        \[
        A_i \leftarrow \frac{R_i^{\mathrm{sum}} - \frac{1}{G} \sum_{j=1}^{G} R_j^{\mathrm{sum}}}{\sqrt{ \frac{1}{G} \sum_{j=1}^{G} \left(R_j^{\mathrm{sum}} - \frac{1}{G} \sum_{k=1}^{G} R_k^{\mathrm{sum}} \right)^2 }}
        \]
              \STATE Compute consistency penalty $A_i^P$ by  Eq.~\ref{con-pen}
      \STATE Apply difficulty-aware weighting:
    \[
      A_i' \gets (A_i - A_i^P) \cdot W(R^{S}_{\text{avg}}),
      \quad \text{where } W(R^{S}_{\text{avg}}) \text{ is calculated by  Eq.~\ref{diff-sample}}
    \]
      \STATE Add sample to buffer:
      \[
        B \gets B \cup \{(q, y_i, A_i')\}_{i=1}^G
      \]
    \ELSE
      \STATE \textbf{continue}
    \ENDIF
    \STATE Update $\pi_\theta$ on buffer $B$ 
  \ENDFOR
\ENDFOR
\STATE \textbf{return} $\pi_\theta$
\end{algorithmic}
\end{algorithm*}

\begin{figure*}[t] 

\centering
\begin{minipage}{0.95\textwidth} 
\begin{promptbox}[\color{black}{Prompt Template for RAG-reasoning Model Generation for Instruction-based Model}]{lightorange}{prompt:selection}

\ttfamily
\quad \quad You are a helpful assistant. Answer the given question.

\quad \quad You must reason **clearly and completely** inside <think> and </think> before providing any final answer.
Always identify and verify all key entities (e.g., person names, locations, dates, awards) mentioned in the question.

\quad \quad If you are uncertain about an entity or fact, or if the question requires external knowledge, you may use <search>your query</search>, and the top search results will be returned between <information> and </information>. Carefully read and reflect on each newly retrieved piece of information.
You can search as many times as you want.

\quad \quad When reasoning, you must ensure your reasoning path aligns strictly with the evidence.

\quad \quad After reasoning, before providing your final answer, rethink it to make sure the answer is exactly correct for the original question.
Use the most accurate span from the evidence when possible.

\quad \quad Only after satisfying all the above, give the final answer inside <answer> and </answer>. For example, <answer>Beijing</answer>.

\quad \quad After outputting the final answer in <answer></answer>, you have one chance to reflect on the answer. If you choose not to reflect, nothing further needs to be done.
Otherwise, you can then re‑examine your thinking process, the information obtained, and even search for more information to verify your previous answer or correct the previous answer. Remember, after reflection ends, you should output the answer in <answer></answer>.
\end{promptbox}
\end{minipage}
\caption{Prompt template for instruction model training.
}
\label{prompt_rag_reasoning_instruct}
\end{figure*}

\begin{figure*}[t] 
\centering
\begin{minipage}{0.95\textwidth} 
\begin{promptbox}[\color{black}{Prompt Template for RAG-reasoning Model Generation for Base Model}]{lightorange}{prompt:base}

\ttfamily
\quad \quad Answer the given question. 

\quad \quad You must reason **clearly and completely** inside <think> and </think> before providing any final answer. 

\quad \quad Always identify and verify all key entities (e.g., person names, locations, dates, awards) mentioned in the question. 

\quad \quad If you are uncertain about an entity or fact, or if the question requires external knowledge, you may use <search>your query</search>, and the top search results will be returned between <information> and </information>. Carefully read and reflect on each newly retrieved piece of information.

\quad \quad You can search as many times as you want. 

\quad \quad When reasoning, you must ensure your reasoning path aligns strictly with the evidence.

\quad \quad After reasoning, before providing your final answer, rethink it to make sure the answer is exactly correct for the original question. 

\quad \quad Use the most accurate span from the evidence when possible. 

\quad \quad Only after satisfying all the above, give the final answer inside <answer> </answer>. For example, <answer> Beijing </answer>. 

\quad \quad After outputting the final answer in <answer> </answer>, you have one chance to reflect on the answer. If you choose not to reflect, nothing further needs to be done. 

\quad \quad Otherwise, you can then re‑examine your thinking process, the information obtained, and even search for more information to verify your previous answer or correct the previous answer. Remember, after reflection ends, you should output the answer in <answer> </answer>. 

\end{promptbox}
\end{minipage}
\caption{Prompt template for base model training.}
\label{prompt_rag_reasoning_base}
\end{figure*}

\begin{figure*}[t]

\centering
\begin{minipage}{1\textwidth}
\begin{promptbox}[\color{black}{Prompt for Sufficient Reward Evaluation}]{lightorange}{prompt:evaluation}
\label{temp:evaluation_prompt}

\ttfamily
You are an expert LLM evaluator that excels at evaluating a QUESTION, ANSWER and REFERENCES. Consider the following criteria: \\
Sufficient Context To The Given Answer: 1 IF the CONTEXT is sufficient to infer the ANSWER to the question and 0  IF the CONTEXT cannot be used to infer the ANSWER to the question. Make the sufficiency judgment based solely on the context, without relying on your memory to determine whether the question can be answered from the context. \\[3pt]
First, output a list of step-by-step questions that would be used to arrive at a label for the criteria. Make sure to include questions about assumptions implicit in the QUESTION. Include questions about any mathematical calculations or arithmetic that would be required. \\[3pt]
Next, answer each of the questions. Please note that you may answer these questions only on the basis of the given context; do not use your own outside knowledge. Make sure to work step by step through any required mathematical calculations or arithmetic. Finally, use these answers to evaluate the criteria. \\[3pt]
EVALUATION (JSON) \\
EXAMPLE: \\
\#\#\# QUESTION \\
In which year did the publisher of Roald Dahl’s Guide to Railway Safety cease to exist? \\
\#\#\# ANSWER \\
2001 \\
\#\#\# References \\
Roald Dahl’s Guide to Railway Safety was published in 1991 by the British Railways Board. The British Railways Board had asked Roald Dahl to write the text of the booklet, and Quentin Blake to illustrate it, to help young people enjoy using the railways safely. The British Railways Board (BRB) was a nationalised industry in the United Kingdom that operated from 1963 to 2001. Until 1997 it was responsible for most railway services in Great Britain, trading under the brand name British Railways and, from 1965, British Rail. It did not operate railways in Northern Ireland, where railways were the responsibility of the Government of Northern Ireland. \\
\#\#\# EXPLANATION \\
The context mentions that Roald Dahl’s Guide to Railway Safety was published by the British Railways Board. It also states that the British Railways Board operated from 1963 to 2001, meaning the year it ceased to exist was 2001. Therefore, the context does provide a precise answer to the question. \\
\#\#\# JSON \\
\{\{"Sufficient Context To The Given Answer": 1\}\} \\[3pt]
Remember the instructions: You are an expert LLM evaluator that excels at evaluating a QUESTION, ANSWER and REFERENCES. Consider the following criteria: \\
Sufficient Context: 1 IF the CONTEXT is sufficient to infer the ANSWER to the question and 0 IF the CONTEXT cannot be used to infer the ANSWER to the question. Make the sufficiency judgment based solely on the context, without relying on your memory to determine whether the question can be answered from the context. \\
First, output a list of step-by-step questions that would be used to arrive at a label for the criteria. Make sure to include questions about assumptions implicit in the QUESTION. Include questions about any mathematical calculations or arithmetic that would be required. \\
Next, answer each of the questions. Please note that you may answer these questions only on the basis of the given context; do not use your own outside knowledge. Make sure to work step by step through any required mathematical calculations or arithmetic. \\
Finally, use these answers to evaluate the criteria. 
Output the \#\#\# EXPLANATION (Text). Then, use the EXPLANATION to output the \#\#\# EVALUATION (JSON)
\end{promptbox}
\end{minipage}
\caption{Prompt used to evaluate context sufficiency.}
\label{prompt_suff_reward}
\end{figure*}
\begin{figure*}[t]
\label{prompt_think_reward}

\centering
\begin{minipage}{0.95\textwidth}
\begin{promptbox}[\color{black}{Prompt for Thinking Reward Evaluation}]{lightorange}{prompt:reasoning-eval}
\label{temp:reasoning_evaluation_prompt}

\ttfamily
You are an expert reasoning evaluator for Retrieval-Augmented Generation (RAG) tasks. \\
Your goal is to judge the reasoning quality of the model's thinking process based on the retrieved context and question. \\
You will assign a reward score between 0 and 1. This score reflects only the quality of the reasoning process, not whether the final answer is correct. \\[3pt]

Evaluation Criteria: \\
1. Logical Soundness – Is the reasoning coherent and structured? \\
2. Contextual Alignment – Does it use retrieved evidence correctly? \\
3. Error Awareness – Does it avoid unsupported assumptions? \\
4. Clarity and Precision – Is it concise, relevant, and non-redundant? \\[3pt]

Scoring: \\
0.0: Completely flawed reasoning \\
1.0: Perfect reasoning \\
Intermediate (e.g., 0.3, 0.7) are allowed. \\[3pt]

Important: \\
- Judge only the thinking process, not the answer. \\
- Reward accurate, grounded, and structured reasoning. \\[3pt]

Your Output: \\
A single float-type score from \{\{0.0, 0.1, 0.2, ..., 1.0\}\}. \\
No explanation. Only the score.
\end{promptbox}
\end{minipage}
\caption{Prompt used to evaluate thinking quality}
\label{prompt_think_reward}

\end{figure*}

\begin{figure*}[t]

\centering
\begin{minipage}{0.95\textwidth}
\begin{promptbox}[\color{black}{Prompt for LLM-As-Judge}]{lightorange}{prompt:answer-verification}
\label{temp:answer_verification_prompt}

\ttfamily
Given a Question and its Golden Answer, verify whether the Predicted Answer is correct. \\
The prediction is correct if it fully aligns with the meaning and key information of the Golden Answer. \\
Respond with True if the prediction is correct and False otherwise. \\[6pt]

Question: \{question\} \\
Golden Answer: \{reference\} \\
Predicted Answer: \{prediction\}
\end{promptbox}
\end{minipage}
\caption{Prompt used for LLM-as-Judge evaluation.}
\label{prompt_llm_as_judge}
\end{figure*}

\begin{figure*}[t]
\centering
\begin{minipage}{0.95\textwidth}

\begin{promptbox}[]{lightblue}{prompt:orange}
\textbf{Question:} Who is Gaius Rubellius Blandus's father-in-law? \\
\textbf{Golden Answer:} Drusus Julius Caesar \\[6pt]

\textbf{Naive GRPO:} 
\texttt{<think>} To find out who Gaius Rubellius Blandus's father-in-law is, I need to first determine who his wife was. I'll search for information about Gaius Rubellius Blandus to find details about his family, particularly his wife. \texttt{</think>}\\
\texttt{<search>} Gaius Rubellius Blandus family \texttt{</search>}\\
\texttt{<information>} ... (information retrieved about Julia Livia being granddaughter of Tiberius, but not focusing on her father) ... \texttt{</information>}\\
\texttt{<think>} From the information retrieved, we know that Gaius Rubellius Blandus married Julia in AD 33. Now, I need to find out who Julia's father was to determine her father-in-law. \texttt{</think>}\\
\texttt{<search>} Julia Livia Gaius Rubellius Blandus wife \texttt{</search>}\\
\texttt{<information>} ... (mentions Julia Livia is granddaughter of Tiberius) ... \texttt{</information>}\\
\texttt{<think>} From the information retrieved, we know that Julia Livia was the granddaughter of Tiberius. Therefore, Gaius Rubellius Blandus's father-in-law is Tiberius. \texttt{</think>}\\
\texttt{<answer>} Tiberius \texttt{</answer>} \\[6pt]

\textbf{TIRESRAG-R1:} 
\texttt{<think>} First, I need to identify who Gaius Rubellius Blandus's wife was. Then I need to find out who her father was. \texttt{</think>}\\
\texttt{<search>} who was Gaius Rubellius Blandus's wife \texttt{</search>}\\
\texttt{<information>} Gaius Rubellius Blandus married Julia Livia, granddaughter of Emperor Tiberius, in AD 33. ... \texttt{</information>}\\
\texttt{<think>} Now that I know his wife is Julia Livia, I need to find out who Julia Livia's father was. \texttt{</think>}\\
\texttt{<search>} who was Julia Livia's father \texttt{</search>}\\
\texttt{<information>} Julia Livia was the daughter of Drusus Julius Caesar and Livilla. ... \texttt{</information>}\\
\texttt{<think>} Therefore, Gaius Rubellius Blandus's father-in-law is Drusus Julius Caesar. \texttt{</think>}\\
\texttt{<answer>} Drusus Julius Caesar \texttt{</answer>}\\
\end{promptbox}

\end{minipage}
\caption{Case study where the naive GRPO model, despite having sufficient information, mistakenly infers the grandfather Tiberius as the father‑in‑law, while TIRESRAG‑R1 correctly reasons through the steps and outputs Drusus Julius Caesar.}
\label{fig:case1}
\end{figure*}

\begin{figure*}[t]
\centering
\begin{minipage}{0.95\textwidth}

\begin{promptbox}[]{lightblue}{prompt:orange}
\textbf{Question:} Where was the place of burial of John De Ferrers, 1st Baron Ferrers of Chartley’s father?\\
\textbf{Golden Answer:} Staffordshire \\[6pt]

\textbf{Naive GRPO:} \texttt{<think>} To find out where John de Ferrers, 1st Baron Ferrers of Chartley’s father was buried, I need to first determine who his father was. I’ll search for information about John de Ferrers, 1st Baron Ferrers of Chartley to find his father’s name. \texttt{</think>}\\
\texttt{<search>} John de Ferrers, 1st Baron Ferrers of Chartley father \texttt{</search>}\\
\texttt{<information>} ... (information retrieved about Roger Ferrers of Fiddington Gloucestershire, unrelated to the correct father) ... \texttt{</information>}\\
\texttt{<think>} From the information retrieved, we know that John de Ferrers, 1st Baron Ferrers of Chartley’s father was Roger Ferrers of Fiddington Gloucestershire. Now, I need to find out where Roger Ferrers of Fiddington Gloucestershire was buried. \texttt{</think>}\\
\texttt{<search>} Roger Ferrers of Fiddington Gloucestershire burial place \texttt{</search>}\\
\texttt{<information>} ... (mentions Fiddington and other unrelated burial info) ... \texttt{</information>}\\
\texttt{<think>} Based on the information retrieved, Roger Ferrers of Fiddington Gloucestershire was buried at Fiddington. Therefore, the place of burial of John de Ferrers, 1st Baron Ferrers of Chartley’s father was Fiddington, Gloucestershire. \texttt{</think>}\\
\texttt{<answer>} Fiddington, Gloucestershire \texttt{</answer>} \\[6pt]

\textbf{TERRISRAG-R1:} \texttt{<think>} First, we need to determine who John de Ferrers, 1st Baron Ferrers of Chartley’s father was. Then we need to find out where his father was buried. \texttt{</think>}\\
\texttt{<search>} John de Ferrers, 1st Baron Ferrers of Chartley father burial place \texttt{</search>}\\
\texttt{<information>} John de Ferrers, 1st Baron Ferrers of Chartley (1271–1324) was the son of Robert de Ferrers, 6th Earl of Derby and Alianore de Bohun. ... \texttt{</information>}\\
\texttt{<think>} From the search results, I can confirm that John de Ferrers, 1st Baron Ferrers of Chartley’s father was Robert de Ferrers, 6th Earl of Derby. Now I need to find where he was buried. \texttt{</think>}\\
\texttt{<search>} Robert de Ferrers, 6th Earl of Derby burial place \texttt{</search>}\\
\texttt{<information>} Robert de Ferrers, 6th Earl of Derby, was buried at St Thomas’s Priory at Stafford, in Staffordshire. ...\texttt{</information>}\\
\texttt{<think>} Therefore, the place of burial of John de Ferrers, 1st Baron Ferrers of Chartley’s father is Stafford, in Staffordshire. \texttt{</think>}\\
\texttt{<answer>} Stafford, in Staffordshire \texttt{</answer>}\\

\end{promptbox}

\end{minipage}
\caption{Case study where the naive GRPO model, despite retrieving sufficient information, follows an incorrect reasoning chain by focusing on an unrelated person (Roger Ferrers of Fiddington) and outputs the wrong burial place. In contrast, TIRESRAG‑R1 reasons through the correct lineage and finds the true location in Staffordshire.}
\label{fig:case2}

\end{figure*}

\begin{figure*}[t]
\centering
\begin{minipage}{0.95\textwidth}

\begin{promptbox}[]{lightblue}{prompt:orange}
\textbf{Question:} Who was born first, Dennis E. Nolan or Humberto Anguiano?\\
\textbf{Golden Answer:} Dennis E. Nolan \\[6pt]

\textbf{Naive GRPO:} \texttt{<think>} To determine who was born first between Dennis E. Nolan and Humberto Anguiano, I need to find out their birth dates. I’ll first search for Dennis E. Nolan’s birth date. \texttt{</think>}\\
\texttt{<search>} Dennis E. Nolan birth date \texttt{</search>}\\
\texttt{<information>} ... Dennis E. Nolan (1872–1956), United States Army general. ... \texttt{</information>}\\
\texttt{<think>} From the search results, I found that Dennis E. Nolan was a United States Army general. Now, I'll search for Humberto Anguiano’s birth date. \texttt{</think>}\\
\texttt{<search>} Humberto Anguiano birth date \texttt{</search>}\\
\texttt{<information>} ... Humberto Anguiano (born 2 November 1910) was a Mexican modern pentathlete. ...\texttt{</information>}\\
\texttt{<think>} Based on the information retrieved, Dennis E. Nolan was born in 1872 while Humberto Anguiano was born in 1910. Therefore, Dennis E. Nolan was born first. \texttt{</think>}\\
\texttt{<answer>} Humberto Anguiano \texttt{</answer>} \\[6pt]

\textbf{TERRISRAG-R1:} \texttt{<think>} Let’s determine who was born first, Dennis E. Nolan or Humberto Anguiano, by finding their respective birth years. \texttt{</think>}\\
\texttt{<search>} Dennis E. Nolan birth date \texttt{</search>}\\
\texttt{<information>} ... Dennis E. Nolan (1872–1956), United States Army general. ... \texttt{</information>}\\
\texttt{<think>} From the search results, I can confirm that Dennis E. Nolan is a United States Army general born in 1872. \texttt{</think>}\\
\texttt{<search>} Humberto Anguiano birth date \texttt{</search>}\\
\texttt{<information>} ... Humberto Anguiano (born 2 November 1910) was a Mexican modern pentathlete. ... \texttt{</information>}\\
\texttt{<think>} Since Dennis E. Nolan was born in 1872 and Humberto Anguiano was born in 1910, Dennis E. Nolan was born first. \texttt{</think>}\\
\texttt{<answer>} Dennis E. Nolan \texttt{</answer>}\\

\end{promptbox}

\end{minipage}
\caption{Case study where the naive GRPO model, despite retrieving the correct birth years (1872 for Dennis E. Nolan and 1910 for Humberto Anguiano), produces an inconsistent reasoning chain and wrongly outputs Humberto Anguiano as born first. In contrast, TIRESRAG‑R1 follows a correct reasoning process and gives the right answer, Dennis E. Nolan.}
\label{fig:case3}
\end{figure*}

\end{document}